%% file: main.tex
\documentclass[3p, 12pt]{elsarticle}

\usepackage{amssymb}
\usepackage{amsmath}
\UseRawInputEncoding 

\usepackage{caption}
\usepackage{subcaption}
\usepackage{multirow}
\usepackage{mathtools}
\usepackage{array}
\usepackage{dirtytalk} 
\usepackage{xcolor}

\usepackage{lineno, hyperref}

\hypersetup{
    colorlinks=true,
    linkcolor=blue,
    filecolor=blue,      
    urlcolor=blue,
    citecolor=blue,
}

\date{}
\journal{}
\makeatletter 
\def\ps@pprintTitle{%
 \let\@oddhead\@empty
 \let\@evenhead\@empty
 \def\@oddfoot{}%
 \let\@evenfoot\@oddfoot}
\makeatother

\newcommand{\revised}[1]{\textcolor{black}{#1}}

\begin{document}

\begin{frontmatter}




\title{Distributed solar generation forecasting using attention-based deep neural networks for cloud movement prediction}

\author[unimelb]{Maneesha Perera\corref{cor1}}
\ead{maneesha.perera1@unimelb.edu.au}
\author[unimelb]{Julian De Hoog}
\ead{julian.dehoog@unimelb.edu.au}
\author[unimelb2]{Kasun Bandara}
\ead{kasun.bandara@unimelb.edu.au}
\author[SLIIT]{Hansani Weeratunge} 
\ead{hansani.w@sliit.lk}
\author[unimelb]{Saman Halgamuge}
\ead{saman@unimelb.edu.au}

\address[unimelb]{Department of Mechanical Engineering, School of Electrical, Mechanical and Infrastructure Engineering, The University of Melbourne, Melbourne, Australia}
\address[SLIIT]{Department of Mechanical Engineering, Sri Lanka Institute of Information Technology, Sri Lanka }
\address[unimelb2]{School of Computing and Information Systems, Melbourne Centre for Data Science, The University of Melbourne, Melbourne, Australia}
\cortext[cor1]{Corresponding author}

\begin{abstract}

Accurate forecasts of distributed solar generation are necessary to maintain grid stability amid the increased uptake of distributed solar photovoltaic (PV) systems. However, the high variability of solar generation over short time intervals (seconds to minutes) caused by cloud movement makes this forecasting task difficult. To address this, using cloud images, which capture the second-to-second changes in cloud cover affecting solar generation, has shown promise. Recently, deep neural networks with \textit{attention} that focus on important regions of an image have been applied with success in many computer vision applications. However, \revised{whether such methods provide meaningful benefits for cloud movement forecasting, and how such improvements propagate through to downstream solar generation forecasting accuracy, remains under-explored.} 
\revised{In this study, we conduct a large-scale empirical investigation of the impact of attention-based cloud forecasting on solar generation forecasting, addressing a gap that has been overlooked in the literature. To this end, we develop a pipeline that incorporates an attention-enhanced convolutional long short-term memory network and an existing self-attention-based video prediction method to forecast cloud movement using satellite imagery. The effectiveness of the resulting cloud forecasts is evaluated through their downstream impact on solar forecasting across 50 PV sites in Australia.}
\revised{We further provide insights into the cloud conditions under which attention-based cloud forecasting methods yield the most significant improvements in downstream solar forecasting accuracy.}
We find that for clouds at high altitudes, the cloud predictions obtained using attention-based methods result in solar forecast skill score improvements of 5.86\% or more compared to non-attention-based methods.

\end{abstract}



\begin{keyword}
Solar photovoltaic power forecasting \sep Deep neural networks \sep Infrared satellite imagery \sep Satellite image forecasting \sep Cloud forecasting 
\end{keyword}

\end{frontmatter}


\section{Introduction}
\label{sec:intro}

The number of small-scale distributed solar photovoltaic (PV) installations on the rooftops of residential and commercial buildings (i.e., behind-the-meter PV systems) has increased over the past decade and is expected to grow exponentially in the coming years \cite{IEA2021World2021, Erdener2022AForecasting}. These PV systems contribute to decarbonising our electricity sector by generating clean and renewable energy. However, integrating large amounts of behind-the-meter PV systems into existing electricity networks poses challenges due to the variable and non-dispatchable nature of solar generation \cite{Erdener2022AForecasting, Hou2019ProbabilisticChina}. Fortunately, accurate predictions of future solar generation (i.e., solar PV power forecasts) can reduce this uncertainty and enable energy stakeholders to better plan and manage electricity grid operation and energy market participation \cite{Yang2019OperationalMarket, Antonanzas2016ReviewForecasting}. In particular, short-term solar power forecasts, ranging from seconds, minutes and up to 1-6 hours ahead, are critical for demand response, scheduling spinning reserves, and maintaining grid stability and power quality \cite{Nespoli2022MachineImagery, Barbieri2017VeryReview, perera2022multi}.

Cloud changes in the sky are the primary driver of solar power fluctuations over short time intervals \cite{Lin2022RecentMethods, Ajith2021DeepData}. Short-term solar forecasting methods typically forecast changes in clouds in two forms: (i) by implicitly forecasting cloud changes using data such as cloud cover measurements provided by nearby weather observation stations, or (ii) by explicitly forecasting cloud movement using images of the sky. The latter enables a more accurate understanding of the movement and formation of clouds, since images of the sky can provide detailed information about clouds at high temporal resolutions (e.g., seconds, minutes) as compared to coarse-grained measures such as cloud cover which is usually available at lower temporal resolutions (e.g., hourly) \cite{Lin2022RecentMethods, Bansal2022ALearning}. 

Imagery of clouds, such as ground-based sky images or satellite images (visible and infrared), are useful data sources to learn changes in clouds that affect solar generation \cite{Ajith2021DeepData, Si2021PhotovoltaicPosition}. Many regions of the earth are now regularly imaged by geostationary satellites (for example, Himawari-8 for Japan and Australia), and this kind of data is becoming a promising source to monitor changes in clouds for behind-the-meter solar PV systems \cite{Si2021PhotovoltaicPosition}. Forecasting solar generation using such satellite images commonly involves a two step learning process \cite{Lin2022RecentMethods}. In the first step, future cloud movement is forecast using past imagery; in the second step, the predicted cloud information is combined with other relevant attributes (e.g., past power generation values, temperature) to forecast solar generation \cite{Bansal2022ALearning, Si2021PhotovoltaicPosition}.  
Much focus in this learning process is given to accurately forecasting cloud movement, since this is the primary driver of short term solar variability \cite{Cheng2022Short-termInterest}. This has also been the main motivation for this work. 


Various methods, such as optical flow \cite{Fu2021SkyForecasting} or particle image velocimetry \cite{Cheng2022Short-termInterest}, have been used to forecast cloud movement in the past.  However, such techniques generally assume that clouds move in a linear manner, and that cloud shape and thickness remain unchanged over time. Deep learning methods, however, can overcome these limitations since they are well suited to learning highly non-linear spatiotemporal relationships in sequential image data. Therefore, more recently deep learning methods have been adapted in the cloud forecasting literature. 

The commonly adapted deep learning methods for cloud forecasting include variants of convolutional neural networks (CNN) or combinations of CNN and long short-term memory networks due to their abilities to capture spatial and temporal relationships in the data. \revised{However, the field of deep learning has seen rapid advances, introducing architectures and components that have demonstrated strong performance across a broad range of 
applications} \cite{VaswaniAshishandShazeerNoamandParmarNikiandUszkoreitJakobandJonesLlionandGomezAidanNandKaiserLukaszandPolosukhin2017AttentionNeed, Huang2017DenselyNetworks, Wang2017PredRNN:LSTMs}.
One such mechanism that has proven successful in many applications is known as \say{attention model} or \say{attention mechanism} \cite{Chaudhari2021AnModels}. The main concept behind attention is giving more importance to useful parts (or areas) of an input to assist the task at hand and focusing less on other areas.
\revised{In the context of solar forecasting, application of attention mechanisms has been investigated in several studies \cite{Qu2021Day-aheadPattern, Lai2021AForecasting, Yu2020ForecastingMemory}. However, these studies primarily apply attention to identify important regions of power generation time-series data rather than to cloud images.} \revised{Few studies have explored applying attention mechanisms in cloud forecasting to identify regions of interests in cloud images. One notable example is \citet{Cheng2022Short-termInterest}, where a CNN-based attention module is used to refine cloud feature maps by weighting satellite image features alongside past power generation values and clear sky predictions, producing a single channel attention-refined feature map for solar power forecasting. While these studies provide valuable insights into how attention mechanisms provide improvements in solar forecasting, several important questions remain underexplored. 
First, the impact of improvements in attention-based cloud forecasting on the accuracy of downstream solar forecasting has not been systematically investigated. Furthermore, given the variety of solar forecasting methods that exist in the literature, it remains unknown whether these improvements are consistent across different forecasting models. Second, there has been limited research on the specific sky conditions in which attention mechanisms offer the most significant benefits.}



\revised{In this paper, we aim to address the above questions. To do so, we take ConvLSTM - a network that laid the foundation for many subsequent architectures proposed for spatiotemporal applications 
\cite{Wang2017PredRNN:LSTMs, Wang2022PredRNN:Learning} as our baseline for cloud movement forecasting. Building on this, we develop an attention-based deep neural network, CBAMConvLSTM, which integrates a convolutional block attention module (CBAM) within ConvLSTM to capture the spatiotemporal evolution of cloud patterns in satellite imagery. We further evaluate SAConvLSTM \cite{Lin2020Self-attentionPrediction}, an 
existing attention-based model originally proposed for video prediction that replaces standard ConvLSTM operations with self-attention, for the task of cloud movement forecasting. The primary motivation of this work is to systematically investigate how improvements in cloud movement forecasting from attention-based deep neural networks translate into improved downstream solar generation forecasting performance. To isolate and attribute forecast improvements specifically to cloud prediction quality rather than architectural differences in the solar forecasting stage, we use well-established solar forecasting networks as downstream models, a multilayer perceptron (MLP), a 1D convolutional neural network (CNN), and a long short-term memory network (LSTM). The cloud forecasts generated by each cloud forecasting model is used as input to each of these downstream networks. We conduct a comprehensive evaluation using a large scale dataset consist of 50 small-scale rooftop solar PV sites in Australia. To further understand in which situations attention provides the greatest benefit, we analyse how cloud conditions corresponding to low and high cloud altitudes (inferred from infrared satellite image pixel values) influence solar forecasting performance.}

In summary, this study provides the following key contributions:

\begin{itemize}

    \item \revised{
        We empirically investigate how attention-based improvements in cloud movement forecasting propagate through to downstream solar generation forecasting. To this end,} 
        \begin{itemize}
        \item \revised{
        We propose CBAMConvLSTM an attention-based extension for ConvLSTM (an existing method widely adapted for spatio-temporal applications) to forecast cloud movement.
        }
        \item \revised{
        We adapt and evaluate an existing self-attention-based network (SAConvLSTM) originally proposed for video prediction to forecast cloud movement.}
    \end{itemize}
    
    \item
    We systematically evaluate how cloud movement forecasts obtained from attention- and non-attention-based networks influence the short-term solar power forecasting performance \revised{using a large scale dataset consisting of 50 small-scale rooftop solar PV sites.}

    \item \revised{We provide novel insights under which cloud altitude conditions (high or low) attention-based cloud forecasting methods provide the greatest benefit to downstream solar generation forecasting accuracy}.
\end{itemize}


The remainder of the paper is structured as follows. Section \ref{sec:relatedwork} discusses the related work; Section \ref{sec:data} presents solar power and satellite image data we use in this work; Section \ref{sec:method} outlines the overall methodology and discusses solar forecasting and cloud forecasting neural networks; Sections \ref{sec:evaluation} and \ref{sec:results} present the evaluation and discuss the results and findings; Section \ref{sec:conclusion} provides the conclusions of this work.

\section{Background and Related Work}
\label{sec:relatedwork}

\say{Solar forecasting} in the literature may refer to forecasting either solar irradiance at a given location or solar PV power generated by a PV system \cite{Erdener2022AForecasting, Lin2022RecentMethods, Antonanzas2016ReviewForecasting, Yang2018HistoryMining, Mayer2021ExtensiveForecasting, BARANCSUK2025122962}. A wider range of forecasting methods such as time series forecasting \cite{Fouilloy2018SolarVariability}, traditional machine learning (e.g., support vector regression or tree-based methods) \cite{Visser2022OperationalDistribution} and neural networks \cite{Kumari2021LongForecasting} have been applied to both irradiance and power forecasting. However, forecasting PV power typically requires either historical power generation data from the system, or system specific information (e.g., panel tilt and orientation) to convert irradiance forecasts to power forecasts.
The input data to any solar forecasting approach can be broadly categorised as data-driven and/or image-based. Data-driven inputs include historical power generation data from the PV site or historical irradiance data, weather observation data, and numerical weather predictions \cite{Lin2022RecentMethods, Mayer2021ExtensiveForecasting, Dolara2015ComparisonPrediction, CloudCast}. Typical image-based inputs include cloud images such as ground-based sky images (GSI) or satellite imagery, which are more commonly used for short-term solar forecasting \cite{Lin2022RecentMethods}.

GSI and satellite images differ based on the techniques used to capture the images and the available temporal and spatial coverage. Satellite images are captured using satellite-sensors while GSI are acquired through digital cameras installed on the ground closer to the PV site \cite{Sobri2018SolarReview, ZHOOLID}. GSI typically offer higher temporal resolutions, ranging from seconds to minutes compared to satellite images that are available at resolutions ranging from minutes to hours. Despite the higher temporal and spatial resolution of GSI, their utility may be limited due to expensive cameras that need to be installed close to a PV site \cite{Bansal2022ALearning}. Furthermore, installing cameras close to hundreds of behind-the-meter PV systems in a neighbourhood or suburb is not practical. As discussed above, geostationary satellites (such as Himawari-8) now provide images for extensive geographical areas and therefore are becoming a promising source of data to monitor changes in clouds \cite{Si2021PhotovoltaicPosition}. Therefore, in this work, we use satellite images to explore cloud movement forecasting. 

The impact of clouds on solar generation is not straightforward as there are several attributes in clouds such as the thickness, altitude (i.e., height from the ground) that can affect the generation from a PV site \cite{HOU2025126243}. For example, \citet{Barbieri2017VeryReview} discussed seven types of clouds that can affect solar generation. Infrared and visible satellite images are useful data sources for inferring cloud attributes such as cloud thickness and altitude levels \cite{Ajith2021DeepData, Si2021PhotovoltaicPosition}. Visible images are captured through sensors that measure the reflected solar radiation from clouds, while infrared images are captured by measuring the heat radiating from the cloud tops. As a result, infrared images can depict cloud cover during both day and night as they are not dependent on the solar radiation. Infrared images are particularly useful to distinguish high vs low altitude clouds while thin vs thick clouds are distinguishable using visible images. While many studies have focused on visible satellite images, limited work has focused on infrared images \cite{Ajith2021DeepData, Cheng2022Short-termInterest} that are useful in inferring some information on cloud altitudes. Despite using infrared imagery, studies in the literature have not investigated the impact of different cloud values inferred from infrared imagery that may correspond to high or low clouds. Thus, through this work, we also intend to shed some light on this direction.

To effectively use cloud attribute information for solar forecasting, it is necessary to accurately model and forecast the spatiotemporal evolution of clouds. Various methods have been used to forecast cloud movement that range from traditional image processing methods (e.g., optical flow method \cite{Fu2021SkyForecasting}, particle image velocimetry \cite{Cheng2022Short-termInterest}) to more recently deep learning methods \cite{Si2021PhotovoltaicPosition}. In the following section, we specifically discuss solar forecasting studies using cloud imagery and deep learning methods that are relevant to the work presented in this paper. 

\subsection{Related Work}

\begin{figure}
    \centering
    \includegraphics[width=\textwidth]{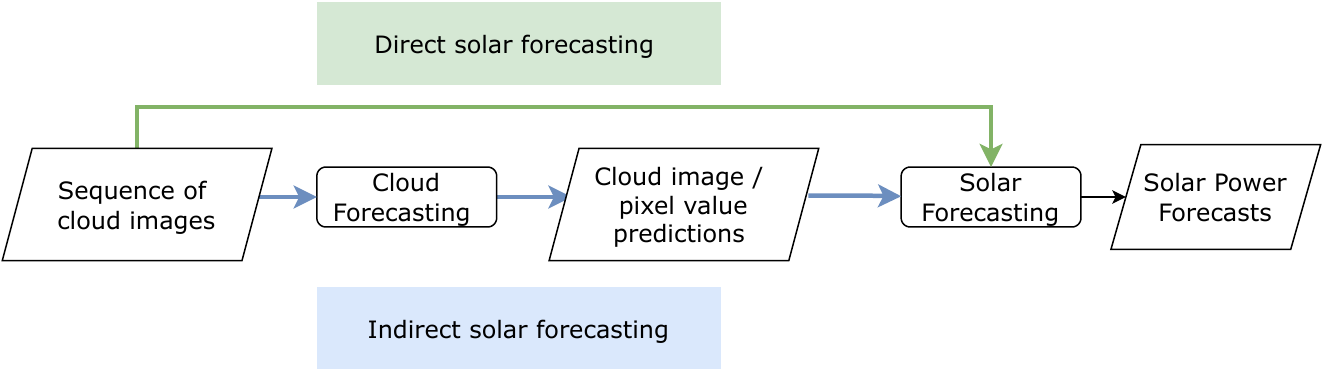}
    \caption{Steps involved for two different approaches (direct vs indirect) of forecasting solar generation using satellite or ground-based sky images. Indirect solar forecasting involves a two step process: forecasting cloud movement and using cloud forecasts as an input to the solar forecasting step. Direct solar forecasting approaches bypass the cloud forecasting step and builds a direct mapping between the cloud images and solar generation.}
    \label{fig:forecasting_process}
\end{figure}

As shown in Figure \ref{fig:forecasting_process}, solar generation forecasting using cloud imagery can be categorized in two main forms (i) indirect forecasting and (ii) direct forecasting. 

The indirect approach involves two steps where the first step involves forecasting the cloud movement. In some studies, the output of this step is a set of pixel values in the image corresponding to the clouds affecting the PV site, while in others the output is a sequence of cloud images where pixels corresponding to the clouds affecting the PV site are extracted afterwards \cite{Si2021PhotovoltaicPosition}. In the second step, the cloud information predicted in the first step is used as an input to forecast solar generation.
In contrast, direct forecasting bypasses the cloud movement forecasting and builds a direct mapping between past cloud images and solar predictions \cite{Zhang2018DeepNowcasting}. Although direct forecasting studies alleviate the two step process they make an assumption that the future solar generation can be explicitly learned using past cloud images. Therefore, in this work, we adapt an indirect solar generation forecasting approach similar to studies \cite{Bansal2022ALearning} and \cite{Si2021PhotovoltaicPosition} to capture cloud dynamics and examine their impact on solar forecasting performance.

Regardless of the approach (indirect or direct), cloud modelling is essentially a spatiotemporal prediction problem. Therefore, some of the commonly adapted deep neural networks for cloud images include variants of convolutional neural networks (CNN) due to their high spatial feature extraction capabilities in images and combinations of CNN and long short-term memory networks (LSTM) due to their ability to model temporal relationships in data. These networks include 2D CNNs, 3D CNNs, CNN + LSTM where the output of the CNN is an input to the LSTM and architectures such as convolutional long short-term memory networks (ConvLSTM) where the long short-term memory network is further modified capture both spatial and temporal dynamics. Table \ref{tab:literature_table} shows some of the solar forecasting studies using cloud images and deep learning.

In \cite{Bansal2022ALearning}, the authors applied a CNN + LSTM to predict three channels of a multispectral satellite image corresponding to a PV system's location given a sequence of past satellite images. The output from the CNN + LSTM, together with historical power generation and temperature data, was used to train a SVR to predict the future solar generation. \citet{Si2021PhotovoltaicPosition} adapted a ConvLSTM (proposed for spatiotemporal applications such as video prediction, precipitation forecasting) to forecast cloud movement and combined the cloud forecasts from ConvLSTM and historical power generation data to forecast solar generation using XGBoost. A study by \cite{Kong2020HybridForecasting} investigated both CNN + LSTM and ConvLSTM. Both networks were used to extract features from GSI combined with a simple multi-layer perceptron network (MLP) to capture features from past PV power data. While both networks demonstrated comparable performance it was observed that ConvLSTM had superior performance over CNN + LSTM. ConvLSTM is one of the earliest deep learning architectures that combined the advantages of convolutional and recurrent networks into a single neural network. ConvLSTM laid the foundation for many subsequent neural network architectures that were proposed for spatiotemporal applications \cite{Shi2015ConvolutionalNowcasting, Wang2022PredRNN:Learning}. Therefore, in our work, we use ConvLSTM as the baseline network to forecast cloud movement.

\begin{table}[!ht]
    \centering
    \begin{tabular}{cp{5cm}p{6cm}}
    \hline
         Literature & Input data &  Method under study \\
    \hline
    \cite{Bansal2022ALearning} & Satellite images, \newline Historical PV power data \newline and temperature data & \textbf{CNN + LSTM} + SVR \\
      \cite{Si2021PhotovoltaicPosition} &  Satellite images, \newline Historical PV power data & \textbf{ConvLSTM} + XGBoost \\
    \cite{Cheng2022Short-termInterest} &  Satellite images, \newline Historical \newline PV power data & \textbf{Auto-encoder with Attention in decoder} \\
    \cite{Zhang2018DeepNowcasting} & GSI and \newline Historical PV power data & \textbf{CNN} + MLP \newline \textbf{CNN + LSTM} + MLP\\
     \cite{Kong2020HybridForecasting} & GSI and \newline Historical PV power data & \textbf{CNN}+ MLP \newline \textbf{CNN + LSTM} + MLP \newline \textbf{ConvLSTM}  + MLP \newline \textbf{PredNet} + MLP \\
     \cite{Ajith2021DeepData} &  GSI, \newline Irradiance data & \textbf{CNN} + LSTM \newline \textbf{CNN + LSTM} + LSTM \\
      \cite{Fu2021SkyForecasting} &  GSI & \textbf{Convolutional Auto-encoders} \\
      \cite{ Zhao20193D-CNN-basedPrediction} &  GSI \newline Irradiance data &
       \textbf{3D CNN} + AR \newline \textbf{3D CNN} + MLP \\
       \cite{wang2025multimodal} &  \revised{Satellite images, Historical PV power data,} \newline \revised{NWP data} &
       \textbf{\revised{ConvGRU}} + \newline \revised{GRU Auto-encoder} \\
    \hline
    \end{tabular}
    \caption{Relevant solar forecasting studies using cloud images and deep neural networks. Deep neural networks applied to process cloud images are marked in boldface under the column Method(s) Under Study.}
    \label{tab:literature_table}
\end{table}

Attention in deep learning is one of the most prominent mechanisms that has boosted the performance of deep neural networks for many applications \cite{Chaudhari2021AnModels, Guo2022AttentionSurvey}. Attention can be applied to different data modalities (e.g., sequential data, images) and multiple forms of attention have been proposed in literature. The intuition behind including any attention mechanism into learning process of a neural network is to focus on informative regions/ parts of the data (i.e., focus on more important features) that can assist with the task at hand \cite{Obeso2022VisualDetection}. This essentially means assigning a higher weight to important features relative to less relevant ones.

Including attention mechanisms has shown promising results in several solar forecasting studies. \citet{Qu2021Day-aheadPattern} applied a simple attention mechanism using two fully connected layers (i.e., dense layers) and softmax activation to forecast solar generation. The attention module was used to weight (i.e., give importance to) short-term and long-term periodicity of previous solar power and temperature time series that can assist future predictions. It was observed that the attention-based method showed superior performance compared to non-attention based methods. Another attention-based method was proposed by \citet{Lai2021AForecasting}, where features extracted from an irradiance time series were weighted using a similar attention module consisting of fully connected layers. The weighted features were passed through a gated recurrent unit network to forecast solar irradiance. When attention mechanisms are applied on image data, they are often referred to as visual attention.

Visual attention mechanisms mimic the human visual system by focusing on relevant regions within an image to make decisions \cite{Guo2022AttentionSurvey}. \citet{Cheng2022Short-termInterest} proposed a direct solar forecasting method using satellite images and visual attention. In their work, the attention module was a CNN that weights satellite image feature maps (i.e., to give high importance to relevant cloud regions) produced by another CNN. The attention module returns a single channel feature map as the output by weighting the satellite image features maps, power at previous time steps and clear sky predictions. These features maps with attention was then used to predict the PV power using a LSTM. It was reported that 1-step ahead forecasts were comparable among attention-based and non-attention based methods, while attention-based method showed improvements for multi-step ahead forecasts. Although, including visual attention has shown a promising direction to improve solar forecasts using cloud images for short-term forecasting, \revised{limited work has investigated how these improvements translate into downstream solar generation forecast performance. As shown in Table \ref{tab:literature_table}, existing studies employ a variety of methods to generate the downstream solar forecast, making it difficult to isolate and attribute performance differences to the quality of the cloud forecast alone. } 


\section{Solar PV Sites and Satellite Data}
\label{sec:data}

\begin{figure}[!ht]
     \centering
     \begin{subfigure}[]{0.3\textwidth}
         \centering
         \includegraphics[width=\textwidth]{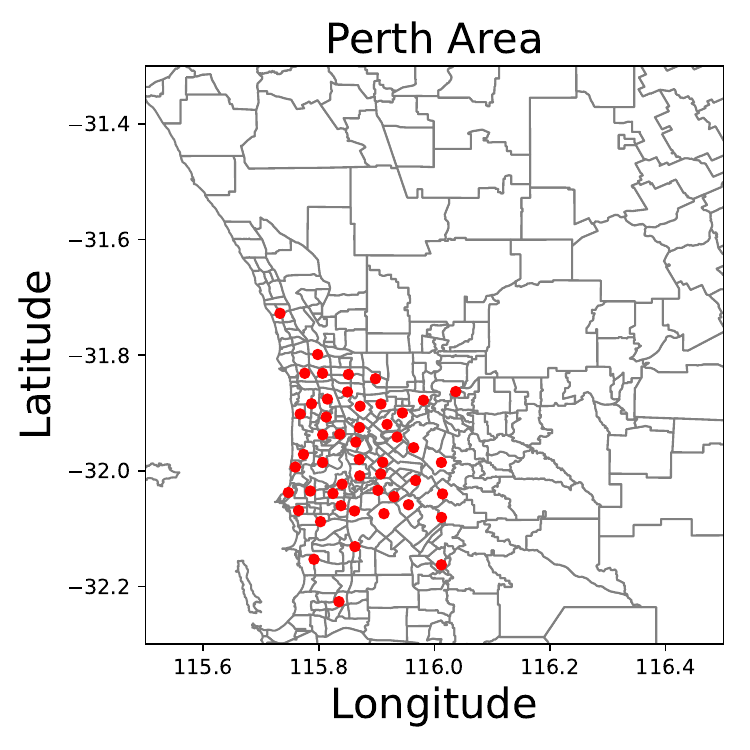}
         \caption{Locations of 50 solar PV sites in Perth marked in red.}
         \label{fig:solar_site_locations}
     \end{subfigure}
     \begin{subfigure}[]{0.3\textwidth}
         \centering
         \includegraphics[width=\textwidth]{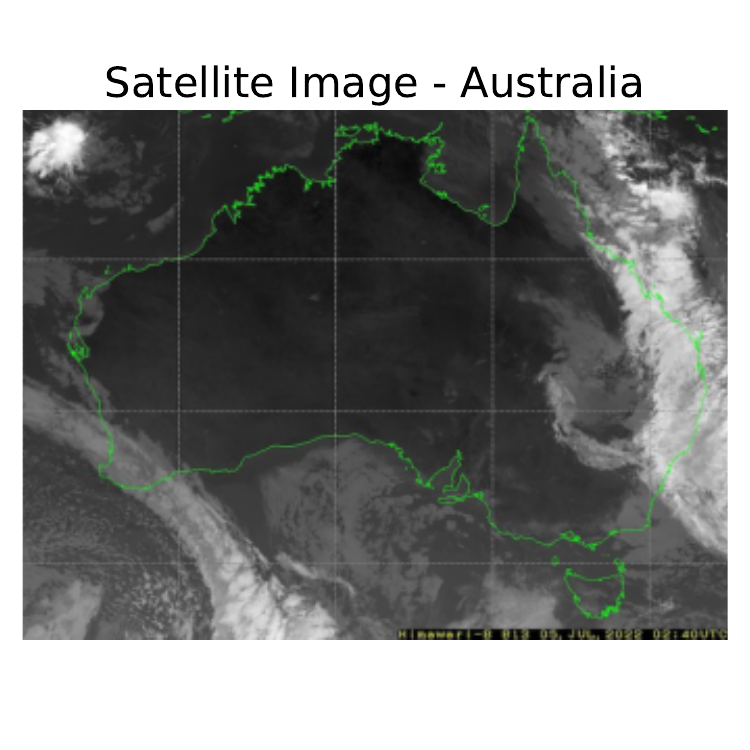}
         \caption{Himawari infrared image for the whole of Australia.}
         \label{fig:himawari_img}
     \end{subfigure}
      \begin{subfigure}[]{0.3\textwidth}
         \centering
         \includegraphics[width=0.9\textwidth]{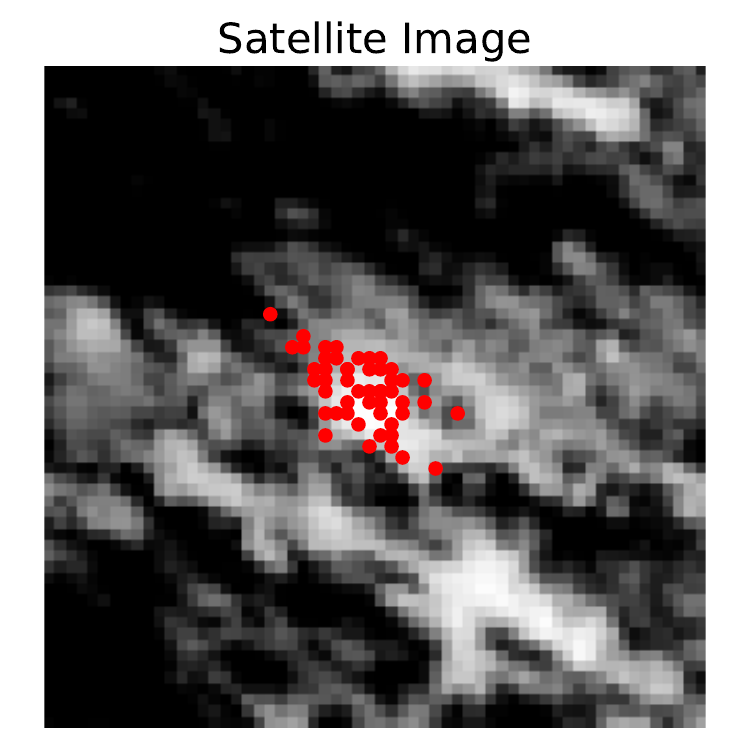}
         \caption{60 X 60 size infrared image covering Perth. Locations of the 50 PV sites are marked in red.}
         \label{fig:perth_SI}
     \end{subfigure}
        \caption{Locations of the 50 PV sites in Perth and infrared satellite image from Himawari.  Locations shown in Figs. \ref{fig:solar_site_locations} and \ref{fig:perth_SI} are in fact the same, but look different due to different projections.}
        \label{fig:solar_sites}
\end{figure}

The solar power generation data we use in this work consists of 50 solar power generation time series obtained from residential and commercial buildings located in Perth, Western Australia. Each solar generation time series consists of one full year of data from February 2020 to February 2021 at 15 minute time intervals. The location of the sites on a map of Perth are shown in Figure \ref{fig:solar_site_locations}. We extract the infrared satellite images from the Meteorological Satellite Center of Japan Meteorological Agency (JMA) website\footnote{https://www.data.jma.go.jp/mscweb/data/himawari/index.html} for the same time period. The satellite images are captured by the Himawari-8 geostationary satellites which takes images of the Earth's surface in 16 spectral bands ranging from 0.47 $\mu m$ to 13.3 $\mu m$. The images we study in this work correspond to the infrared channel (band 13) ranging from 10.4 $\mu m$ to 11.2 $\mu m$. The satellite images have a spatial resolution of 2 km (i.e., each pixel in the image covers an area of 2 km $\times$ 2km) and are available at 10 minutes time intervals. Therefore, we re-sampled the solar generation time series to 10 minute intervals to ensure consistency with the satellite images. Figure \ref{fig:himawari_img} shows an infrared satellite image for entire Australia from the JMA website. In our work, we extract a $60 \times 60$ pixel area from the satellite images such that the center of the image corresponds to the latitude and longitude of the locations of PV sites in Perth as shown in Figure \ref{fig:perth_SI}. As the spatial resolution of the satellite image is within few kilometers, the location of a PV site may be represented using a single pixel in the image \cite{Si2021PhotovoltaicPosition, Bansal2022ALearning}. Therefore, the pixel value in the satellite image reflects the clouds that are vertically above the PV site of interest.

\begin{figure}[!ht]
     \centering
     \begin{subfigure}[]{\textwidth}
         \centering
         \includegraphics[width=\textwidth]{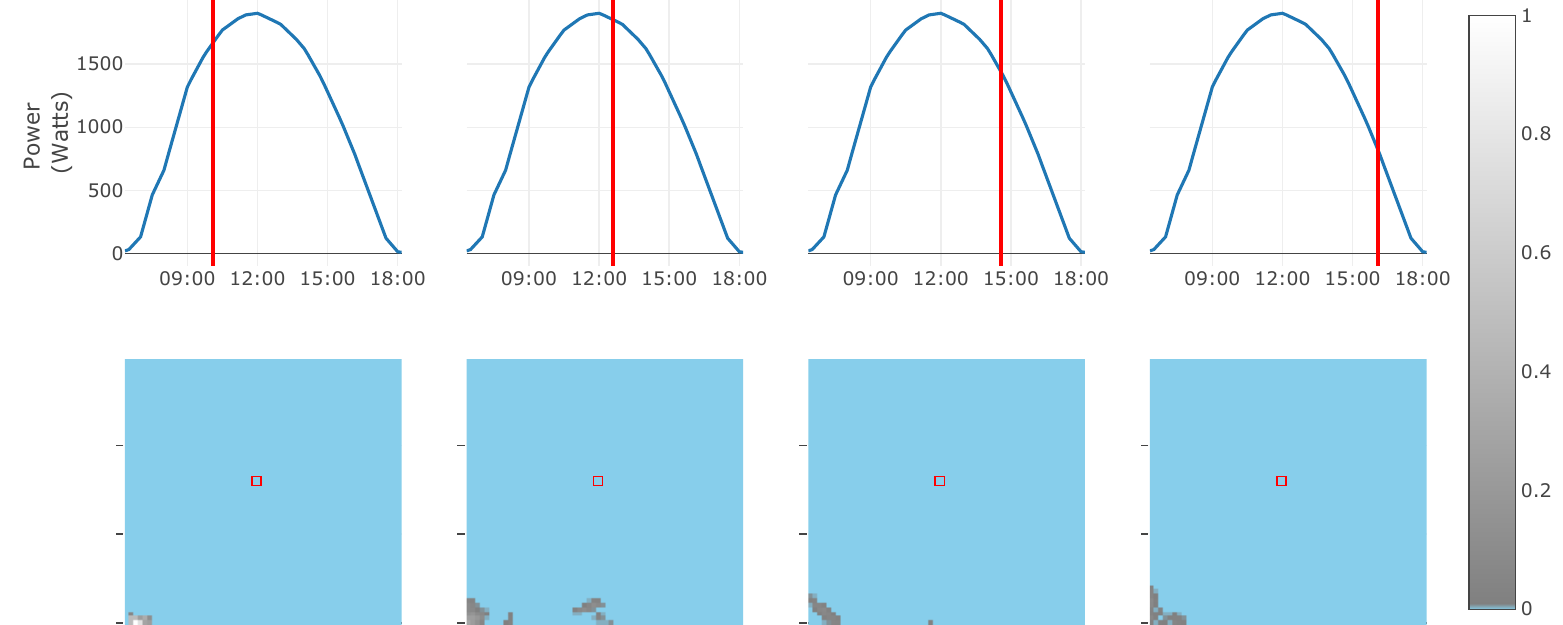}
         \caption{Solar power generation time series from a PV site and satellite image on a clear sky day.}
         \label{fig:sunny_day}
     \end{subfigure}
     \par\bigskip
     \begin{subfigure}[]{\textwidth}
         \centering
         \includegraphics[width=\textwidth]{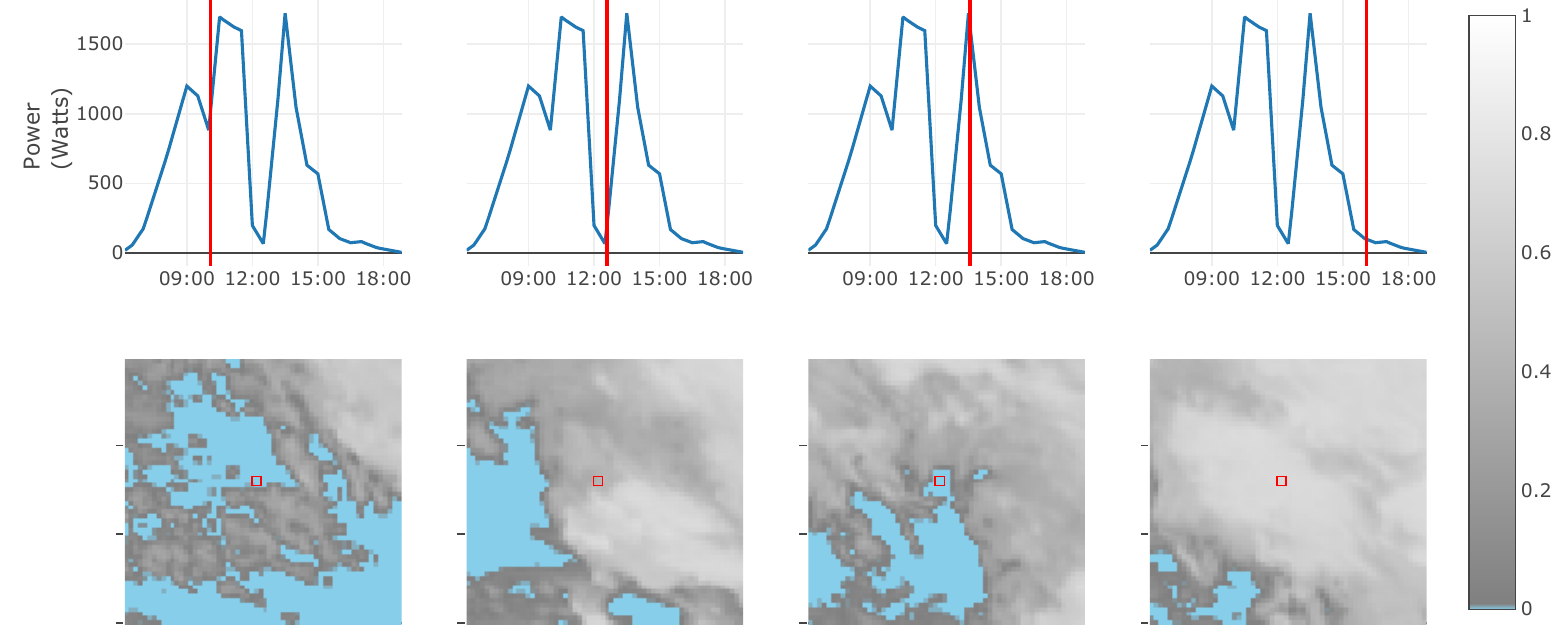}
         \caption{Solar power generation time series from a PV site and satellite image on a cloudy day.}
         \label{fig:cloudy_day}
     \end{subfigure}
        \caption{Solar power generation time series from a PV site and infrared satellite images corresponding to the time marked in red in the PV power generation time series. The location of the PV site in the satellite image is marked in a red rectangle which indicates the cloud condition vertically above the PV site. Areas without clouds are shown in blue color. Color scale of the satellite images refer to the pixel value that correlates to altitude levels of clouds (1- high, 0-low).}
        \label{fig:generation_SI}
\end{figure}

As can be seen in Figure \ref{fig:himawari_img}, the infrared satellite image is a grayscale image where each pixel value shows an integer between 0 and 255 (0-black, 255- white). As we discussed before, clouds are identified in infrared images using temperature sensors that measure the heat radiating from the cloud tops. Since the temperature decreases with increasing height of the troposphere, clouds that reside at higher altitudes (such as \textit{Cirrus-cirrostratus} clouds) have cold tops, while lower altitude clouds (such as \textit{Cumulus}, \textit{Stratocumulus} clouds) have warmer tops \cite{Barbieri2017VeryReview}. Clouds that have very cold tops are shown in bright white in the infrared images, while clouds with warmer tops will not appear as white in the image. Therefore, infrared images can be used to infer information about cloud altitudes, where high altitude clouds may appear closer to white colour, while low altitude clouds may not appear very white. Moreover, black shows that there are no clouds present (i.e., clear sky). 

Figure \ref{fig:generation_SI} shows the solar power generation time series of one PV site and the corresponding satellite image (where the location of the PV site in the satellite image is marked with a red rectangle) for the timestamp marked in red in the time series. To better represent the different cloud altitude levels that may be present in the satellite image, we use a color scale to show the pixel values of the original grayscale image. Figure \ref{fig:sunny_day} shows a day where the solar power generation from the PV site has a complete power generation profile without any fluctuations during the day. As can be seen in the corresponding satellite image, there are no clouds covering the PV site and hence the power generation is not interrupted. However, as shown in Figure \ref{fig:cloudy_day}, where the power generation is interrupted during the middle of the day at multiple time intervals, the corresponding satellite image shows clouds covering the location of the PV site.

\begin{figure}[!ht]
    \centering
    \includegraphics[width=\textwidth]{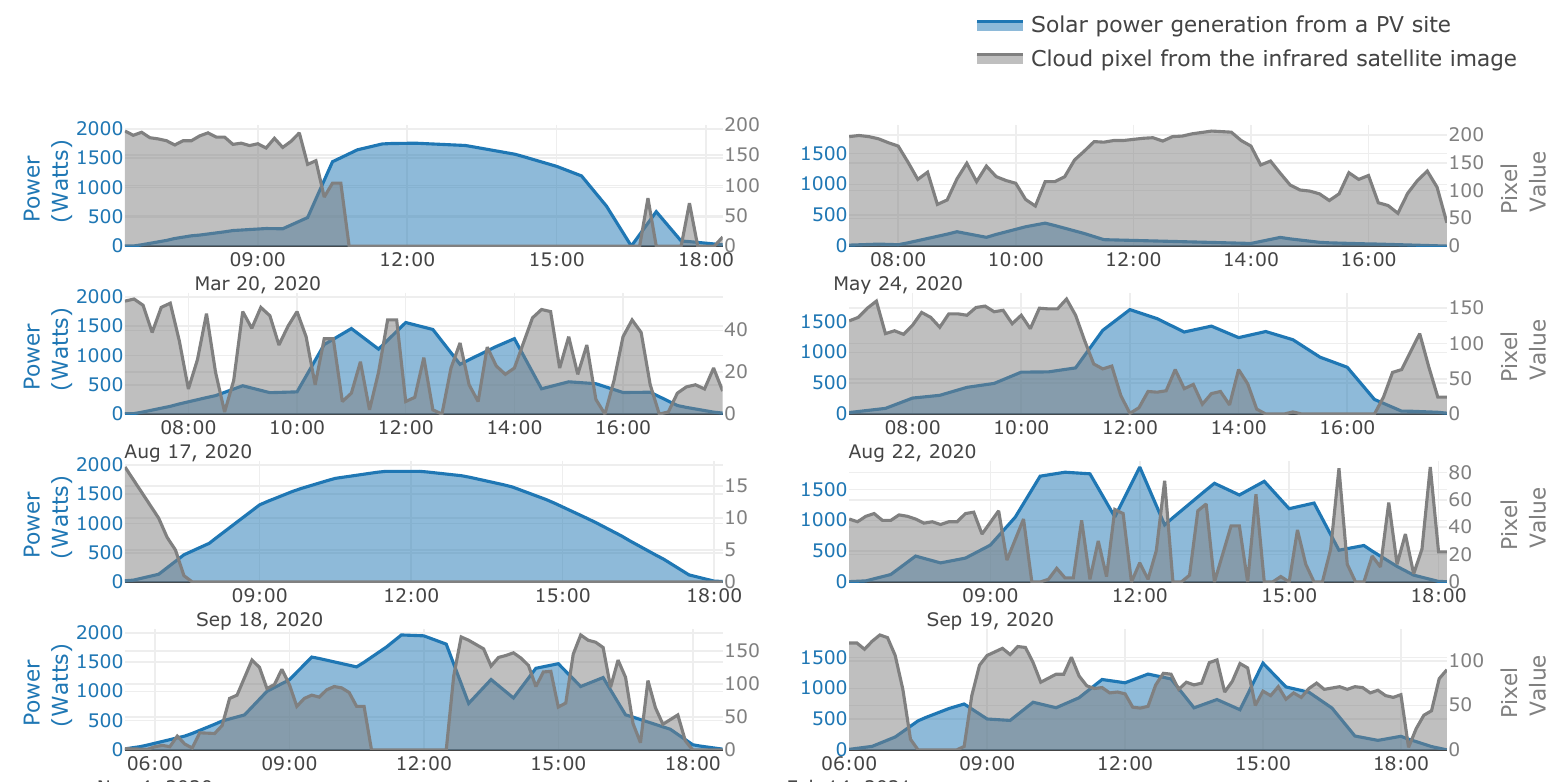}
    \caption{Solar power generation of a PV site in Perth on eight different days and the pixel value of the infrared satellite image corresponding to the cloud condition above the site.}
    \label{fig:cloud_pixel_solar_generation}
\end{figure}

Similar observations can be made by examining Figure \ref{fig:cloud_pixel_solar_generation} which shows the solar power generation for one PV site and the pixel value from the satellite image showing the altitude of the cloud vertically above the PV site on eight different days. We can see that in most scenarios (e.g., Mar 20 2020 between 12:00 - 16:00, Sept 18 2020, Nov 4 2020 around 12:00) when the pixel value is $0$ (no clouds or clear sky) the solar power generation from the site is not interrupted. In contrast, during periods where the pixel value is not $0$, indicating the presence of clouds, fluctuations of the power generation are visible (e.g., Mar 20 2020 between 8:00-11:00, May 24 2020, Sept 19 2020). However, a clear correlation between the solar power generation and cloud pixel (i.e., the cloud altitude) cannot be derived solely through visual observations of the data where high altitude may have a lower impact on the solar generation (due to the scattering effect of the sunlight) and low altitude clouds may have a high impact.

\section{Methodology}
\label{sec:method}

\subsection{Overall Framework}
\label{sec:framework}

\begin{figure}[!ht]
    \centering
    \includegraphics[width=\textwidth]{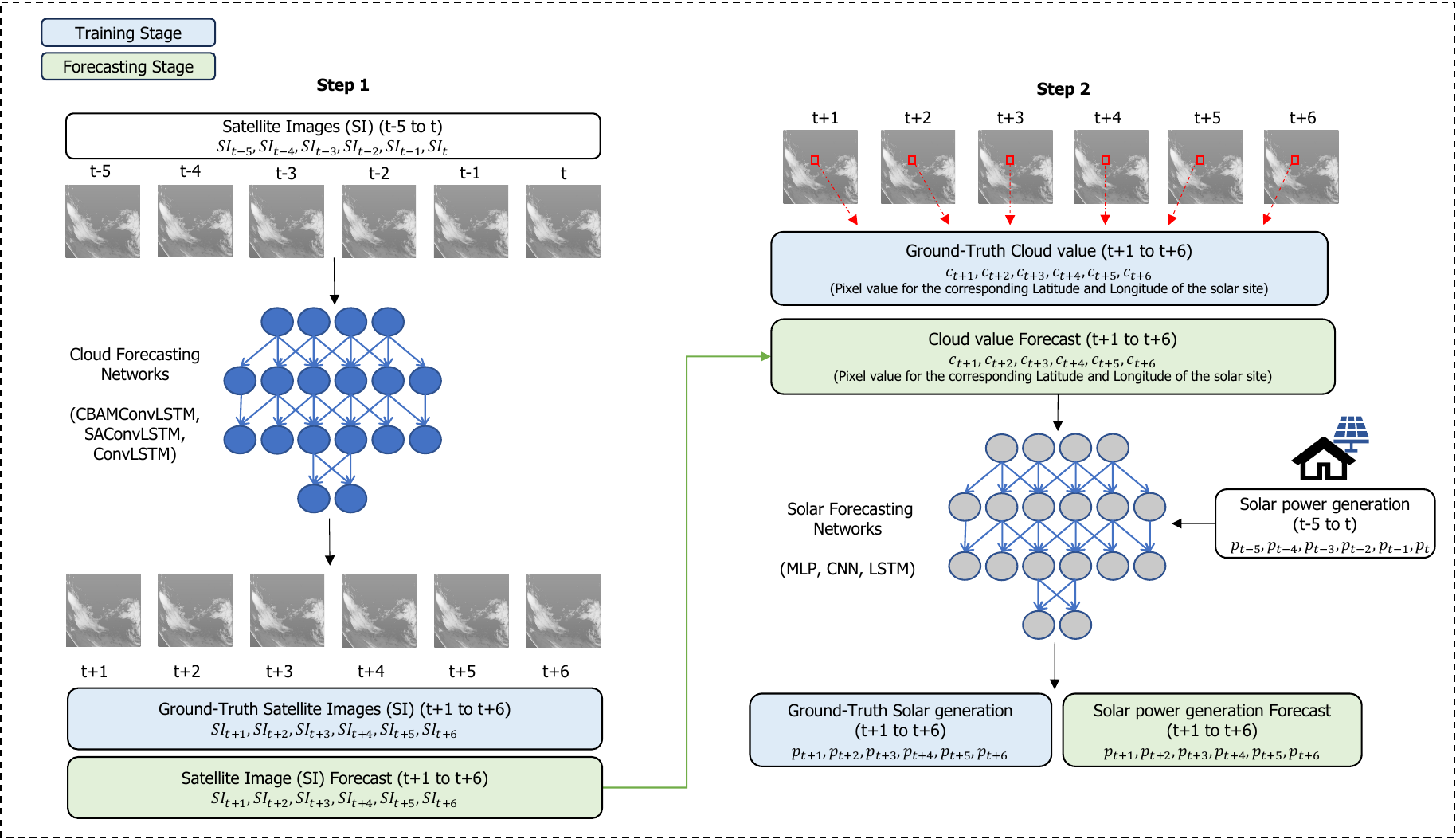}
    \caption{\revised{Solar power forecasting framework to study the impact of cloud movement predictions from attention and non-attention-based networks on downstream solar forecasting performance.}}
    \label{fig:overall_framework}
\end{figure}

Figure \ref{fig:overall_framework} shows the overall framework of our work which consists of two stages (i) Training Stage, (ii) Forecasting Stage. We consider a short-term forecast horizon of 1 hour at 10 minute time intervals (i.e., 6 time steps into the future) where the primary driver of solar power fluctuations during these time intervals are cloud movements \cite{Lin2022RecentMethods}.


Both the \textit{Training Stage} and the \textit{Forecasting Stage} (i.e., testing phase) involve training and forecasting two categories of networks \revised{(for cloud forecasting and solar forecasting)}, as shown in Figure \ref{fig:overall_framework}. \revised{During the Training Stage the cloud forecasting and solar forecasting networks are trained separately. However, during the Forecasting Stage the output of the cloud forecasting networks are used as an input to the solar forecasting networks.}
\begin{enumerate}
    \item \textbf{Cloud movement forecasting:}  
    \begin{enumerate}
         \item Training Stage:  To forecast cloud movement, we train the deep neural networks discussed in section \ref{sec:cloud_forecasting_dnn}. The inputs to the cloud forecasting networks include 6 satellite images of size $60 \times 60$ corresponding to the clouds in the past hour as shown in Figure \ref{fig:overall_framework}. Since a $60 \times 60$ satellite image covers the locations of all PV sites under study, we train a single cloud forecasting network (of the same type) to forecast the movement of clouds that affect all 50 PV sites. 
         \item Forecasting Stage: We obtain the infrared satellite image predictions from the trained cloud forecasting network for the forecasting hour (shown as step 1 in Figure \ref{fig:overall_framework}).
     \end{enumerate}
    
    \item \textbf{Solar generation forecasting:} 
     \begin{enumerate}
         \item Training Stage:  We train multiple neural networks discussed in section \ref{sec:solar_forecasting_dnn} to forecast the power generation at each of the PV sites. For each PV site, we train a separate network of the same type where the inputs to the network include power generation values of the past hour and ground-truth cloud pixel values (corresponding to the PV site extracted from the infrared satellite image) for the prediction time period ($t+1, \dots ,t+6$) -- as shown in Figure \ref{fig:overall_framework}.
         
         \item Forecasting Stage: As shown in step 2 in Figure \ref{fig:overall_framework}, using the outputs from the cloud movement forecasting (from above), together with the power generation values of the past hour as inputs, we generate solar forecasts using the trained solar forecasting network corresponding to the PV site.
     \end{enumerate}

\end{enumerate}


\subsection{Deep Neural Networks for Cloud Forecasting}
\label{sec:cloud_forecasting_dnn}

Forecasting cloud movement is a spatiotemporal prediction problem where the time or temporal factor relates to the changes observed across continuous satellite images and the space or spatial factor relates to the formations of clouds within a satellite image \cite{Si2021PhotovoltaicPosition}. Several deep neural networks have been proposed for spatiotemporal learning in the deep learning literature \cite{Wang2022PredRNN:Learning}. One of the widely adapted structures is the combination of CNN and LSTMs, which combines the advantages of convolutional and recurrent neural networks. \textit{Convolutional LSTM} (i.e., ConvLSTM) proposed by \citet{Shi2015ConvolutionalNowcasting} is one of the very early deep neural networks that is widely adapted in spatiotemporal applications and has also been adapted to forecast cloud movement in multiple solar forecasting studies \cite{Si2021PhotovoltaicPosition, Paletta2021BenchmarkingAnalysis, Kong2020HybridForecasting}.

In ConvLSTM, the convolutional and recurrent neural networks are combined together by replacing the matrix multiplication within a standard LSTM cell with the convolution operation to capture spatial dependencies in addition to the long and short-term modelling of the LSTM. Equation \ref{eq:subeq1}-\ref{eq:subeq4} shows the operations of a ConvLSTM and the matrix multiplications replaced with convolution operations are marked in boldface. $i_t$, $o_t$, $f_t$, $C_t$ and $h_t$ represent the input gate, output gate, forget gate, cell state and hidden state at time $t$. $W_{<gate/cell>, <input/hidden>}$ are the weight matrices for the respective gate or cell state associated with the hidden state or input state and $b_{gate/cell}$ indicates the corresponding bias vectors.

\begin{subequations}
\begin{align}
    i_t = \sigma(W_{ih} \textbf{*} h_{t-1} + W_{ix} \textbf{*} x_{t} + b_{i})  \label{eq:subeq1} \\
    o_t = \sigma(W_{oh} \textbf{*} h_{t-1} + W_{ox} \textbf{*} x_{t} + b_{o}) \label{eq:subeq2}\\
    f_t = \sigma(W_{fh} \textbf{*} h_{t-1} + W_{fx} \textbf{*} x_{t} + b_{f}) \label{eq:subeq3}\\
    C_t = i_t \odot \tanh(W_{ch} \textbf{*} h_{t-1} + W_{cx} \textbf{*} x_{t} + b_{c}) + f_{t} \odot C_{t-1} \\
    h_{t} = o_{t} \odot \tanh(C_t) \label{eq:subeq4}
\end{align}
\label{eq:lstm}
\end{subequations}

Compared to the traditional LSTM that ignores spatial information, ConvLSTM can extract spatial dependencies and is therefore suitable for spatiotemporal applications where both spatial and temporal dynamics of the data need to be extracted. We train a ConvLSTM to forecast the satellite image sequence of the next hour in 10 minute time intervals (i.e., to predict 6 images). We use this network as a benchmark method for cloud forecasting to evaluate the attention-based networks we investigate in this work.

\subsubsection{Attention-based Deep Neural Networks}

Attention model or attention mechanism in neural networks in a simple form refers to giving attention (i.e., weight) to important parts of an input that can assist the task at hand and ignore (or give less focus) to the other parts and thereby enhancing the ability of a neural network to process and learn from data \cite{Chaudhari2021AnModels}.
\newline
\newline
\newline
\textbf{CBAM ConvLSTM}
\vspace{0.3cm}
\newline
In this paper, we integrate an attention mechanism into the standard ConvLSTM structure to improve the ability of ConvLSTM to process spatiotemporal information by focusing on more important parts of an input image. We specifically consider the task of predicting a sequence of infrared satellite images where our objective by adding an attention mechanism to the ConvLSTM is to focus and learn the cloud regions of the infrared image, in contrast to the non-cloud regions. We adapt an attention module proposed in \cite{Woo2018CBAM:Module} known as convolutional block attention module (i.e., CBAM). CBAM is an attention module that can be seamlessly integrated into any convolutional neural network architecture to refine intermediate feature maps generated through a convolution layer by adding attention to these feature maps.

\begin{figure}[!ht]
    \centering
    \includegraphics[width=0.7\textwidth]{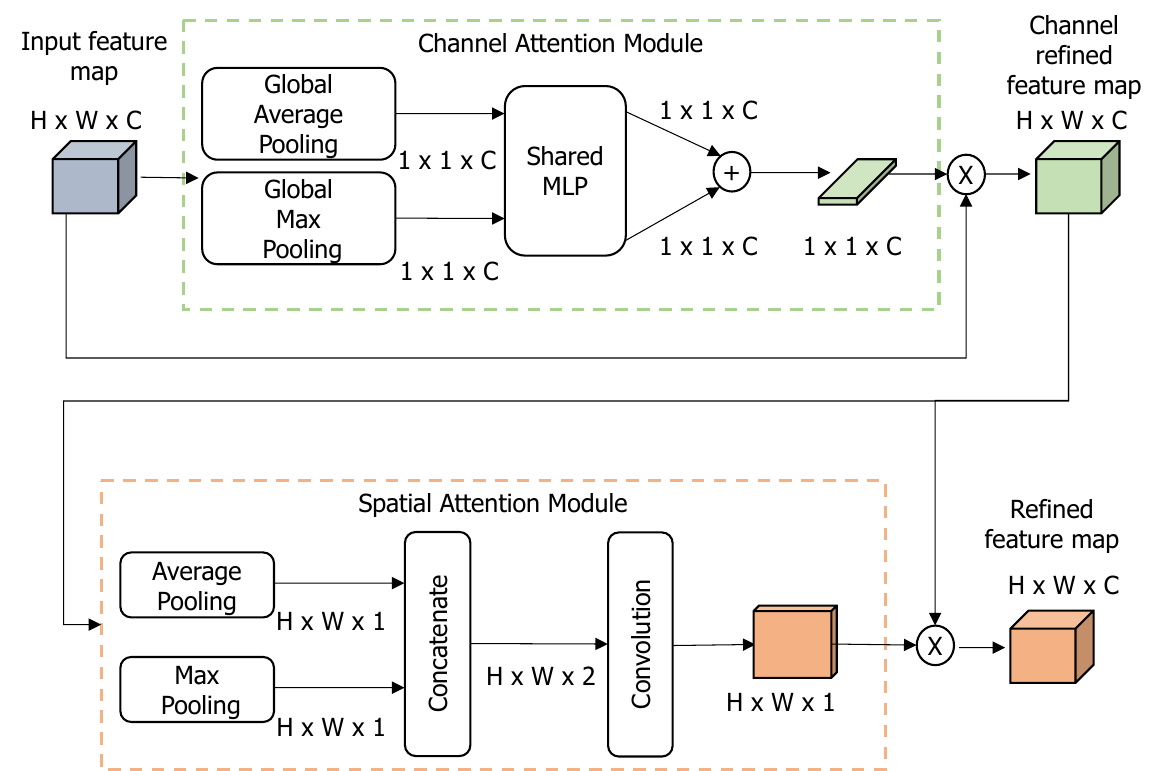}
    \caption{Basic structure of the Convolutional Block Attention Module (CBAM). $H, W, C$ indicates the size of three different dimensions. $\otimes$ shows element wise multiplication and $+$ indicates addition.}
    \label{fig:cbam}
\end{figure}

The basic structure of the CBAM is shown in Figure \ref{fig:cbam}. The attention module focuses on adding attention to two dimensions - the channel and spatial dimension. The channel attention module focuses on what to focus on in the image (i.e., what is meaningful). In contrast, the spatial attention module is responsible for identifying where to focus (i.e., where is the informative part in the image). 

As can be seen in Figure \ref{fig:cbam}, first, the input feature map from a convolution layer is passed to the channel attention module. In this module, to focus on the channel dimension, first, the spatial information of an image is aggregated through global average and max pooling. In global average/ max pooling the average/ maximum across the spatial dimension is derived to convert a $H \times W \times C$ dimensional input to $1 \times 1 \times C$ dimensional output where $H$, $W$, $C$ indicates the size of each dimension. Afterwards, these down-sampled feature maps are passed through a shared MLP to produce a refined feature map across the channel dimension. The final channel refined feature map is obtained by multiplying the original feature map with the attention map obtained from the channel attention module. 

The channel refined feature map is then passed through the spatial attention module, where at first, the feature map is down-samples across the channel dimension through average and max pooling. Afterwards, these two feature maps are combined together, and a convolution operation is applied to obtain an attention map across the spatial dimension. Finally, the spatial attention map is multiplied with the channel refined attention map to derive the final refined feature map with attention across both channel and spatial dimensions. 

\begin{figure}[!ht]
    \centering
    \includegraphics[width=0.7\textwidth]{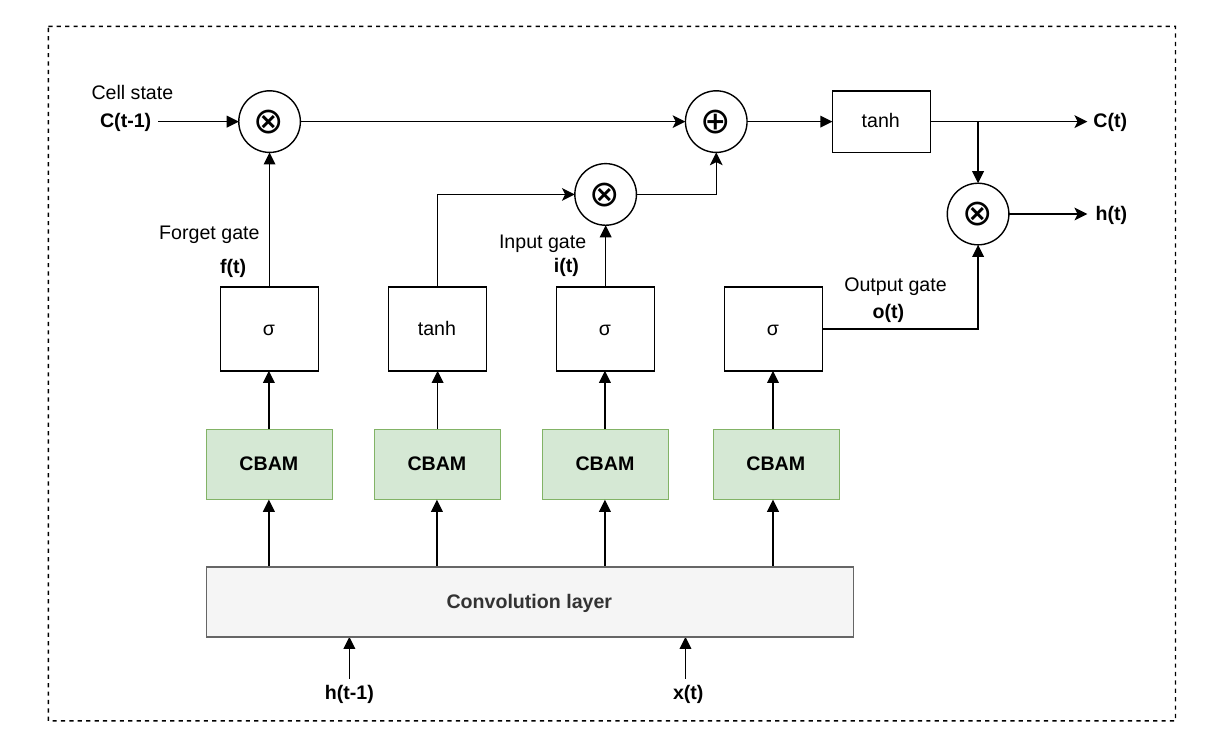}
    \caption{\revised{CBAM ConvLSTM architecture. The addition of the CBAM modules to the ConvLSTM architecture is shown in green colour. $\otimes$ shows element wise multiplication and $+$ indicates addition.}}
    \label{fig:convlstm-cbam}
\end{figure}

In this work, we integrate a CBAM-based attention mechanism to the ConvLSTM structure after each convolution operation within the ConvLSTM to refine (through attention) the intermediate feature maps that are generated after each convolution operation. \revised{Figure \ref{fig:convlstm-cbam} shows the integration of the CBAM module to the current ConvLSTM architecture.} Equations \ref{eq:subeq1-cbam} - \ref{eq:subeq4-cbam} show the operations of CBAMConvLSTM where the changes to the standard ConvLSTM operations by adding the CBAM are shown in boldface.

\begin{subequations}
\begin{align}
    i_t = \sigma(\textbf{CBAM(}W_{ih} * h_{t-1}\textbf{)} + \textbf{CBAM(}W_{ix} * x_{t}\textbf{)} + b_{i})  \label{eq:subeq1-cbam} \\
    o_t = \sigma(\textbf{CBAM(}W_{oh} * h_{t-1}\textbf{)} + \textbf{CBAM(}W_{ox} * x_{t}\textbf{)} + b_{o}) \label{eq:subeq2-cbam}\\
    f_t = \sigma(\textbf{CBAM(}W_{fh} * h_{t-1}\textbf{)} +\textbf{CBAM(} W_{fx} * x_{t}\textbf{)} + b_{f}) \label{eq:subeq3-cbam}\\
    C_t = i_t \odot \tanh(\textbf{CBAM(}W_{ch} * h_{t-1}\textbf{)} + \textbf{CBAM(}W_{cx} * x_{t}\textbf{)} + b_{c}) + f_{t} \odot C_{t-1} \\
    h_{t} = o_{t} \odot \tanh(C_t) \label{eq:subeq4-cbam}
\end{align}
\label{eq:cbam_lstm}
\end{subequations}
\newline
\newline
\textbf{SAConvLSTM}
\vspace{0.3cm}
\newline
In this work, we further adapt a self-attention based ConvLSTM architecture (SAConvLSTM) that was proposed by \citet{Lin2020Self-attentionPrediction} for video prediction tasks to forecast cloud movement using infrared satellite images. SAConvLSTM relies on an attention mechanism known as self-attention (also refereed to as intra-attention) where the main idea is to capture global dependencies in an input sequence \cite{VaswaniAshishandShazeerNoamandParmarNikiandUszkoreitJakobandJonesLlionandGomezAidanNandKaiserLukaszandPolosukhin2017AttentionNeed}. In the context of images, self-attention mechanism calculates a correlation among pixels (i.e., pairwise relations) in an image to capture global spatial dependencies \cite{Lin2020Self-attentionPrediction}. 

SAConvLSTM embeds the self-attention mechanism to the standard ConvLSTM to capture global spatial dependencies and further introduces an additional memory module to remember global spatial dependencies captured in past time steps when predicting an image sequence. To relate this concept to the context of cloud forecasting in a simplified form, capturing global dependencies in an image refers to identifying how each pixel (showing clouds in a particular location) relates to every other pixel (clouds in another location) in the image, and memorizing global spatial dependencies refers to remembering relevant spatial features captured from the past cloud images in addition to the spatial features of the current cloud image. To enable capturing such long-range spatial dependencies, SAConvLSTM has a memory module named self-attention memory module that is embedded to the standard ConvLSTM. 

The following equations show the operations within the addition memory module which takes the input hidden state $h_{t}$ of the current time step calculated from standard ConvLSTM operations and $M_{t-1}$ (newly introduced memory) from the last time step. Furthermore to memorize global dependencies an aggregated feature map $Z$ is generated by applying self-attention on the hidden state $h_t$ and $M_{t-1}$ features.  The following equations shows the calculations of newly introduced memory $M$ in \cite{Lin2020Self-attentionPrediction}.

\begin{subequations}
\begin{align}
    i'_t = \sigma(W_{i'h} * h_t + W_{i'z} * Z + b_{i'})  \label{eq:subeq1-sa_convlstm} \\
    o'_t = \sigma(W_{o'h} * h_{t} + W_{o'z} * Z + b_{o'}) \label{eq:subeq2-sa_convlstm}\\
    M_t = i'_t \odot \tanh(W_{mh} * h_t + W_{mz} * Z + b_{m}) + (1 - i'_{t}) \odot M_{t-1} \label{eq:subeq4-sa_convlstm} \\
    \hat{h_{t}} = o'_{t} \odot \tanh(M_t) \label{eq:subeq3-sa_convlstm} 
\end{align}
\label{eq:sa_convlstm}
\end{subequations}

Similar to the ConvLSTM, we train the attention-based ConvLSTMs (both \textit{CBAMConvLSTM}, \textit{SAConvLSTM}) to forecast the satellite images for the next hour (6 time steps ahead). During the forecasting stage, for all cloud forecasting networks, we conduct an autoregressive prediction to forecast the next 6 satellite images from $t+1$ to $t+6$. In this approach the predicted cloud images from $t+1$ to $t+(k-1)$ are given as inputs to predict $t+k$ image for all $k>1$. 

We train the networks using the \textit{Adam} optimizer to minimize the structural similarity index measure (SSIM) loss which takes into account luminance (i.e., brightness), contrast and structure. SSIM is more suitable to measure the similarity of the images compared to the mean squared error loss that only measures the difference between corresponding pixels of the ground-truth and predicted image \cite{Wang2004ImageSimilarity}. Using the SSIM loss enables the network to learn to generate images structurally similar to the ground-truth images. 

Due to the computational complexity of training cloud forecasting networks using multiple hyperparameters, we limit the hyperparameter search to a fixed set of values. In particular, we tune the hyperparameters shown in Table \ref{tab:hyperparameters_dnn} using a grid search approach where a fixed number of values to explore for each hyperparameter is given. Moreover, to ensure the network does not over-fit to the training images, we train all networks for 200 epochs with early stopping to stop training if there is no improvement in the validation loss in $10$ consecutive epochs. At the beginning of training, the learning rate of all networks is set to $0.001$, and a learning rate scheduler is used to reduce the learning rate by a factor of $0.1$ if there is no improvement in $5$ consecutive epochs. 

\begin{table}[!ht]
    \centering
    \begin{tabular}{cc}
    \hline
        Hyperparameter & Range \\ 
    \hline
         number of layers & 1, 2, 3, 4, 5, 6 \\
         number of hidden states per layer & 32, 64 \\ 
         batch size & 16, 32 \\
    \hline
    \end{tabular}
    \caption{Hyperparameters and parameter ranges provided for cloud forecasting networks.}
    \label{tab:hyperparameters_dnn}
\end{table}

\subsection{Solar Forecasting Neural Networks}
\label{sec:solar_forecasting_dnn}

To forecast solar generation and to evaluate the impact of cloud predictions on their solar forecasting performance, we use three widely used state-of-the-art neural networks in the solar forecasting literature \cite{perera2024day, Castangia2021AForecasting}: Multi-Layer Perceptron (MLP), 1D Convolutional Neural Network (CNN), Long Short-Term Memory Network (LSTM).

The first network is a \textit{MLP} which is a simple feed forward neural network architecture. In feed forward neural networks the information is only passed in one direction from input neurons of the network to the output neurons. The information is passed through several hidden layers of the neural network in a MLP structure to learn a non-linear mapping between the input and output \cite{Svozil1997IntroductionNetworks}. As we discussed above MLP is adapted in several forecasting studies in the literature to forecast solar generation and is one the common neural networks used to compare against more advanced networks \cite{Fouilloy2018SolarVariability, duPlessis2021Short-termBehaviour, Castangia2021AForecasting}. 

The second network is a \textit{LSTM} and is a type of recurrent neural network architecture that consists of feedback loops to carry information from one input step to the next in contrast to traditional feed forward neural networks. Feedback loops in RNN cells enables to capture temporal patterns in data allowing the information flow from one time step to the other and therefore, is adapted in many sequence prediction problems including forecasting studies \cite{Hewamalage2021RecurrentDirections}. LSTM cells have an additional gating mechanism that allows to control the flow of information by preserving relevant information and forgetting irrelevant information which enables LSTMs to capture both long and short term temporal dynamics in the data compared to traditional RNNs. 

The third network we use for solar power forecasting is a \textit{1D CNN}. CNNs are a class of deep neural networks that are translation-invariant, powerful feature extractors widely adapted in computer vision applications \cite{Liu2017AApplications}. The filter (i.e., kernel) that is essentially a weight vector is the main component in a CNN and it slides through the input and learns to recognise specific patterns in the data. Depending on the dimensions of the input and the features need to be extracted different filter dimension are used. For example 2-dimensional filters are more commonly used to capture feature across the height and width dimensions of an image. For 1-dimensional data such as time series, relationships among data points exist in the temporal dimension and therefore 1-dimensional filters are more commonly applied \cite{Mellit2021DeepForecasting}. 

In this work, as discussed above, we train a separate neural network: MLP, LSTM, CNN for each PV site and train these networks to minimize the mean squared error loss with the \textit{Adam} optimizer. Moreover, we tune the hyperparameters shown in Table \ref{tab:hyperparameters} for each of the networks. For hyperparameter tuning we use an optimization method \textit{Bayesian optimization} from Hyperopt python package \cite{Bergstra2013MakingArchitectures}. Compared to other hyperparameter tuning methods such as grid search that conduct an exhaustive search to find the optimal huperparameters, bayesian optimization method limits the number of exhaustive evaluations by using information from prior evaluations and is therefore much more efficient to tune a multiple hyperparameter combinations with large ranges of possible values.

\begin{table}[!ht]
    \centering
    \begin{tabular}{clr}
    \hline
        Network & Hyperparameter & Range \\ 
    \hline
        MLP &  learning rate & 0.0001 - 0.1\\
         &  number of hidden layers & 1 - 5\\
         &  number of neurons per layer & 1 - 256\\
         &  number of epochs & 100 - 2000\\
         &  learning rate & 0.0001 - 0.1\\
         &  batch size & 1 - size of training samples\\
         &  drop out rate & 0 - 0.5\\
        1D CNN & learning rate & 0.0001 - 0.1 \\
        & kernal/filter size & 2, 3 \\
        & number of filters & 32, 64 \\
        & number of convolution layers & 1 - 5 \\
        & drop out rate & 0 - 0.5 \\
        & number of epochs & 500 - 2000 \\
        & batch size & 1 - size of training samples \\
        LSTM & learning rate & 0.0001 - 0.1\\
         &  number of lstm layers & 1 - 5\\
         &  number of neurons per layer & 32, 64\\
         &  number of epochs & 500 - 2000\\
         &  learning rate & 0.0001 - 0.1\\
         &  batch size & 1 - size of training samples\\
    \hline
    \end{tabular}
    \caption{Hyperparameters tuned in this work and the hyperparameters ranges provided to find the optimal hyperparameters for each neural network.}
    \label{tab:hyperparameters}
\end{table}

\section{Evaluation}
\label{sec:evaluation}

During the \textit{Forecasting Stage}, we evaluate the impact of cloud forecasts produced by the cloud forecasting networks under the following three benchmark scenarios.


\begin{enumerate}
    \item \textit{Ground truth clouds }- The first scenario uses cloud pixels from the ground-truth satellite images as an input to the solar forecasting networks. This condition shows the best or \say{ideal} solar power forecasts that can be achieved as the cloud conditions in the future time instants are known. 
    \item  \textit{Persistence clouds }- The second scenario uses the cloud image observed at last timestamp before the start of the forecasting hour as inputs to the solar forecasting network without predicting the satellite images. This condition acts as a benchmark to determine if the image predictions from the cloud forecasting networks have improved over simply using the last observed image for the prediction time period. 
    \item \textit{No clouds included-} The third scenario includes training the same solar forecasting networks discussed in section \ref{sec:solar_forecasting_dnn} with power generation data from the past hour as the sole input to the networks. This condition is used to determine if including cloud values derived from the infrared satellite images have an impact on forecasting solar power generation.
\end{enumerate}

Table \ref{tab:tarin_val_test} shows the number of training, validation and testing samples of the solar power forecasting networks per PV site. For the three cloud forecasting networks, we use similar numbers of training, validation and test image samples.

\begin{table}[!ht]
    \centering
    \begin{tabular}{cr}
    \hline
         Stage & Number of samples \\ 
    \hline
         Training & 16,915 \\
         Validation & 4229 \\ 
         Testing/ Forecasting & 2289 \\
    \hline
    \end{tabular}
    \caption{Number of training, validation and test samples.}
    \label{tab:tarin_val_test}
\end{table}

To evaluate the solar power forecasting performance, we use two error measures - Root Mean Squared Error Skill Score (RMSE Skill Score) and Mean Absolute Error Skill Score (MAE Skill Score) commonly used in the literature \cite{Voyant2017MachineReview, Yang2020VerificationForecasts}. RMSE Skill Score and MAE Skill Score measure the quality of forecasts compared to a reference model, and therefore provide better error interpretability and comparison across data with different scales. The most commonly used reference model is the persistence method where the forecasts are similar to the last observed power generation value \cite{Bansal2022ALearning}. The following shows the calculation of the above error metrics where $h$ indicates the forecast horizon (6 steps in our case) and $\hat{P}_i$, $P_i$ refers to the predicted and actual power generation values at a given time $i$.

\begin{subequations}
\begin{align}
    RMSE  = \sqrt{\frac{1}{h}\sum_{i=t+1}^{t+h}{(\hat{P}_i -P_i)}^2} \\
    RMSE \; Skill \; Score = \left ( 1 - \frac{RMSE_{forecasting \; method}}{RMSE_{persistence}}\right ) * 100\% 
\end{align}
\label{eq:metrics-rmse}
\end{subequations}

\begin{subequations}
\begin{align}
    MAE = \frac{1}{h}\sum_{i=t+1}^{t+h}|\hat{P_i} -P_i|\\
    MAE \; Skill \; Score = \left ( 1 - \frac{MAE_{forecasting \; method}}{MAE_{persistence}} \right ) * 100\% 
\end{align}
\label{eq:metrics-mae}
\end{subequations}

A higher value for the RMSE/ MAE skill score indicates that the forecasting method performs better than a persistence method. A skill score of 0\% indicates the forecasts quality is similar to the quality of a persistence method. We calculate the RMSE and MAE skill scores across all test samples of a PV site and then take the average error across all test samples of a PV site. Finally, we calculate the average RMSE/ MAE skill scores across all 50 PV sites obtain the final errors.

\section{Results and Discussion}
\label{sec:results}

\subsection{Cloud Forecasts - Visualisations}

\begin{figure}[!ht]
     \centering
     \begin{subfigure}[]{0.45\textwidth}
         \centering
         \includegraphics[width=\textwidth]{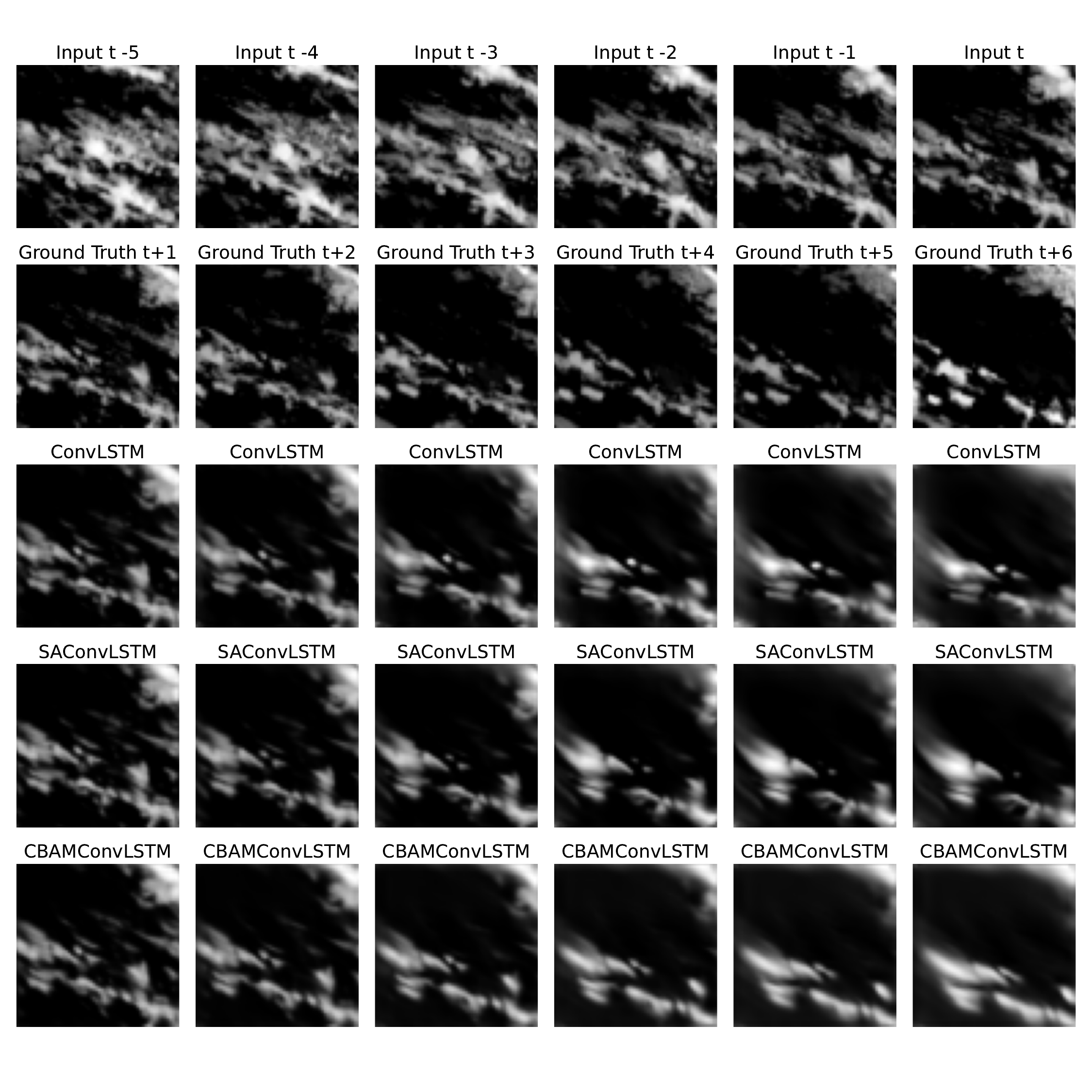}
         \caption{Test Sample 1}
         \label{fig:cloud_predictions_1}
     \end{subfigure}
     \begin{subfigure}[]{0.45\textwidth}
         \centering
         \includegraphics[width=\textwidth]{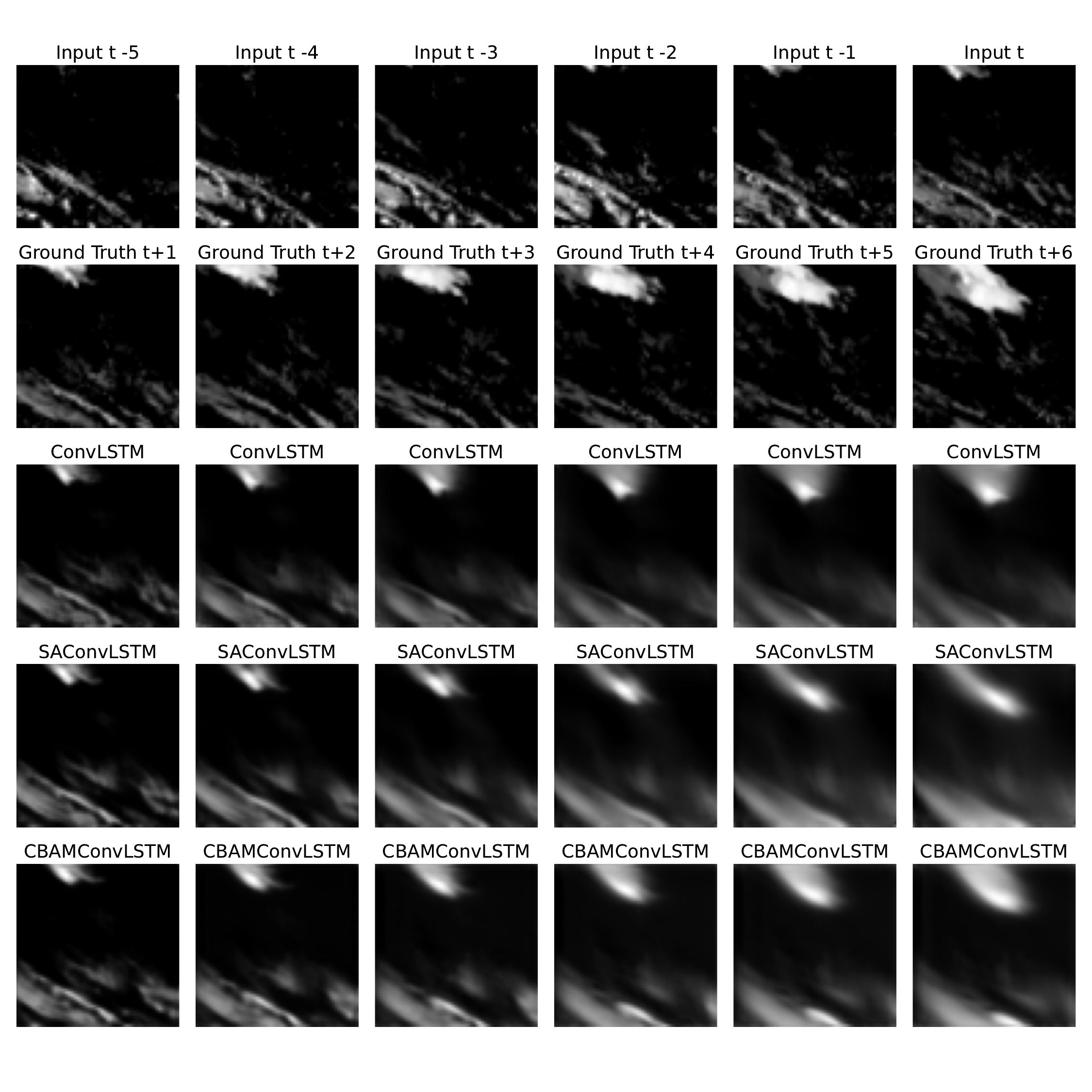}
         \caption{Test Sample 2}
         \label{fig:cloud_predictions_2}
     \end{subfigure}
     \hfill
      \begin{subfigure}[]{0.45\textwidth}
         \centering
         \includegraphics[width=\textwidth]{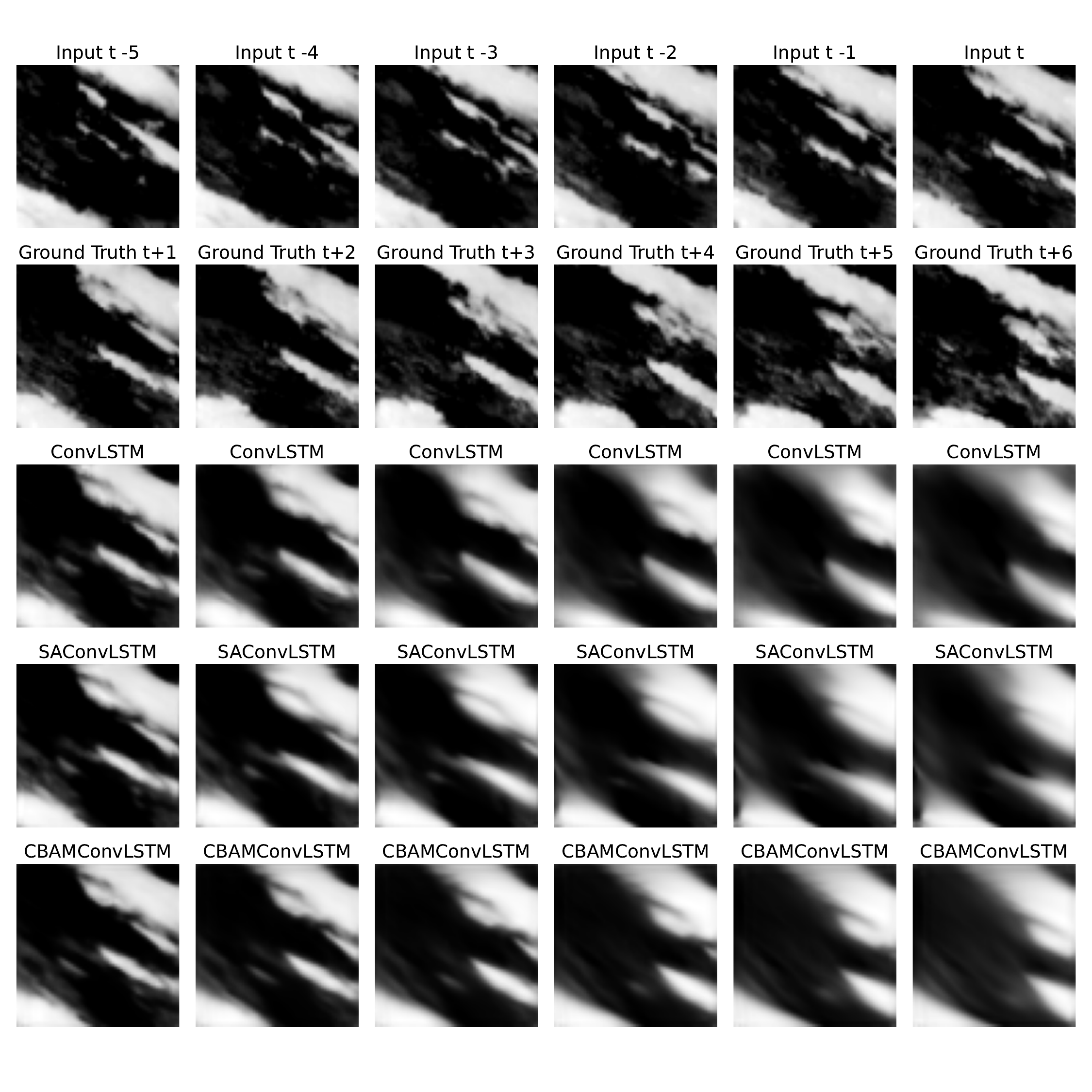}
         \caption{Test Sample 3}
         \label{fig:cloud_predictions_5}
     \end{subfigure}
      \begin{subfigure}[]{0.45\textwidth}
         \centering
         \includegraphics[width=\textwidth]{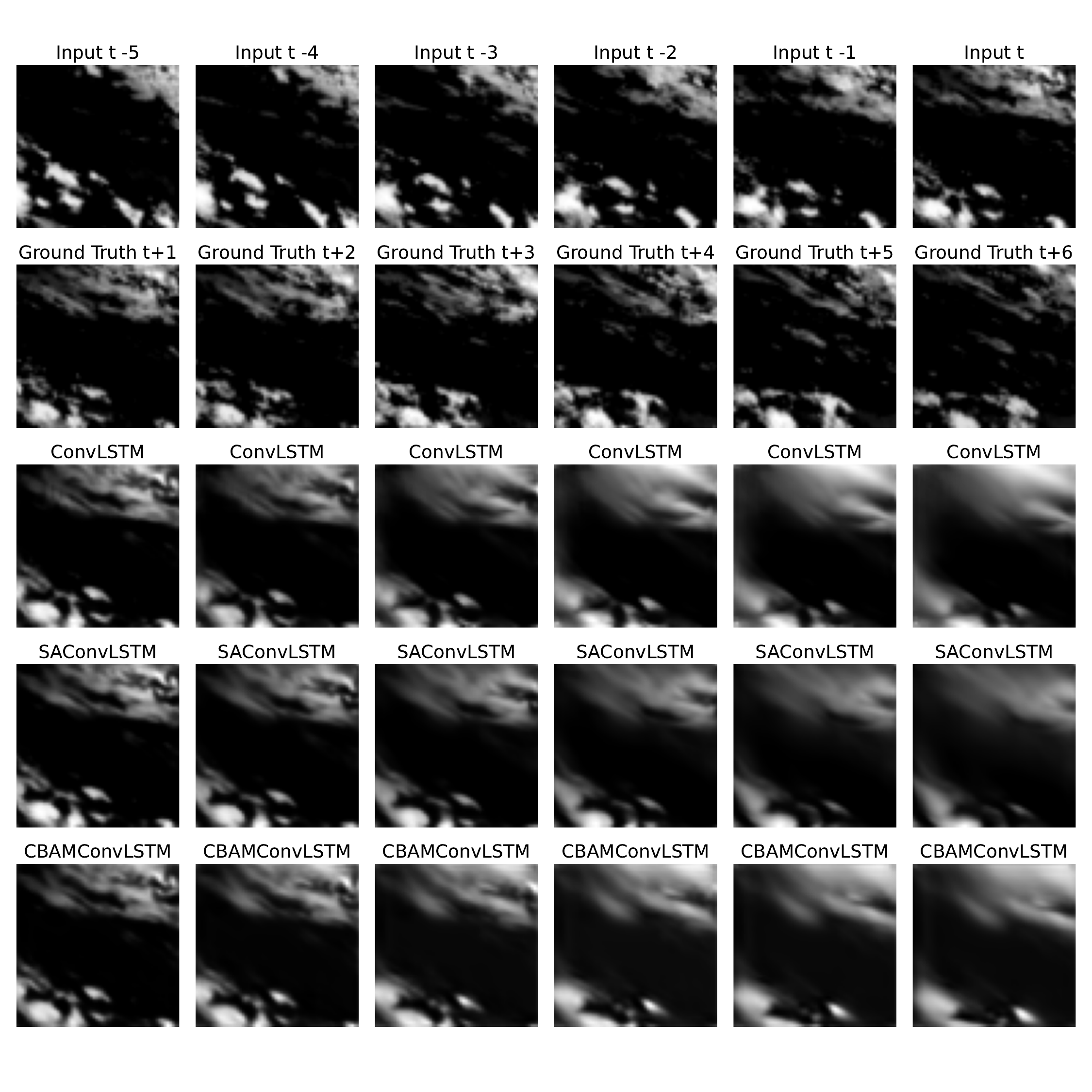}
         \caption{Test Sample 4}
         \label{fig:cloud_predictions_4}
     \end{subfigure}
        \caption{Visualisation examples of predicted cloud images from ConvLSTM, SAConvLSTM and CBAMConvLSTM for 4 randomly selected test samples with varying cloud conditions. The first row shows the input images to the network, the second row shows the ground truth satellite images for the forecasting period and the next three rows shows the predictions from ConvLSTM, SAConvLSTM and CBAMConvLSTM networks respectively.}
        \label{fig:cloud_predictions}
\end{figure}

Figure \ref{fig:cloud_predictions} shows the predictions from cloud forecasting deep neural networks for four randomly selected test samples with varying cloud conditions. We can observe that all cloud forecasting networks produce output images similar to the ground truth image in the immediate future time instant ($t+1$). However, we can see that as the number of time instants to the future increase the similarity with the ground truth image is reduced. We quantitatively evaluate the quality of these cloud forecasts and their impact on quality of the solar power forecasts considering all PV sites under study in section \ref{sec:results_quantitative}. \revised{
As the solar forecasting network parameters are fixed after training and the only varying input is the source of the cloud forecast, any observed differences in solar forecast accuracy can be directly attributed to the quality of the cloud predictions, and therefore provides a clear and interpretable indicator of cloud forecast quality.}

\subsection{Impact of Cloud Forecasts on Solar Power Forecasts}
\label{sec:results_quantitative}

Table \ref{sec:results} shows the average RMSE Skill Score across the 50 PV sites considering all test samples for different cloud inputs during the forecasting stage. As discussed above, \say{Ground-truth clouds} shows the best solar forecasts (i.e., \say{ideal} scenario) that can be achieved as the ground-truth cloud values are known. Therefore, if the cloud forecasts are similar to the ground-truth, the skill score will be closer to the ground-truth clouds scenario. The RMSE Skill Score difference in comparison to using ground-truth clouds is also shown in the Table. A lower value in the Skill Score Difference column therefore, indicates the solar forecasts are closer to the best skill score that can be achieved. MAE skill scores are presented in \ref{appendixA}.

\begin{table}[!ht]
    \centering
    \begin{tabular}{p{6cm}>{\raggedleft\arraybackslash}p{1cm}>{\raggedleft\arraybackslash}p{1.2cm}>{\raggedleft\arraybackslash}p{1cm}>{\raggedleft\arraybackslash}p{1.2cm}>{\raggedleft\arraybackslash}p{1cm}>{\raggedleft\arraybackslash}p{1.2cm}}
    \hline
   Cloud Input During  & \multicolumn{6}{c}{Solar Power Forecasting Network}\\
    Forecasting Stage &   \multicolumn{2}{c}{MLP} & \multicolumn{2}{c}{CNN} & \multicolumn{2}{c}{LSTM} \\
    \cline{2-7}
    & \footnotesize \textit{RMSE Skill Score (\%)} & \footnotesize \textit{Skill \newline Score \newline Difference} & \footnotesize \textit{RMSE Skill Score (\%)} & \footnotesize \textit{Skill \newline Score \newline Difference} & \footnotesize \textit{RMSE Skill Score (\%)} & \footnotesize \textit{Skill \newline Score \newline Difference} \\
    \hline
    Ground-truth clouds & 34.26 & - & 32.79 & - & 40.28 & - \\
    Forecasted clouds from \newline ConvLSTM & 30.00 & 4.26 & 28.85 & 3.94 & 35.46 & 4.82 \\ 
    Forecasted clouds from \newline SAConvLSTM & 31.33 & 2.93 & 30.02 & 2.77 & 36.82 & 3.46\\ 
    Forecasted clouds from \newline CBAMConvLSTM & 31.57 & \textbf{2.69}  & 30.39 & 2.40 & 37.15 & \textbf{3.13}\\ 
    Persistence clouds & 28.63 & 5.63 & 27.02 & 5.77 & 34.02 & 6.26 \\ 
    No clouds included & 18.27 & 15.99 & 30.87 & \textbf{1.92} & 36.60 & 3.68\\
    \hline
    \end{tabular}
    \caption{Average RMSE Skill Score across 50 rooftop solar PV sites considering all test samples of a site and the respective skill score difference with regard to using \textit{Ground-truth clouds} showing the best skill score that can be achieved. The lowest skill score difference is marked in boldface.}
    \label{tab:results}
\end{table}

When comparing the ground-truth clouds and no clouds scenarios, we can see that including clouds as an input can improve the solar forecasts. However, the LSTM trained without clouds can achieve a RMSE skill score of $36.60 \%$, which is greater than MLP and CNN networks using ground-truth clouds as inputs. This may relate to the ability of LSTMs to capture long-term temporal patterns of time series data compared to other networks and, therefore, have been able to forecast the next-hour by effectively learning temporal patterns in the past hour. Nonetheless, we can see that including clouds to the LSTM network further improve the solar forecasts.

Considering the scenarios where cloud forecasts are given as an input, we can see that irrespective of the type of solar forecasting network, including cloud forecasts from ConvLSTM, SAConvLSTM and CBAMConvLSTM can improve the forecasts compared to using persistence cloud forecasts as an input. Furthermore, cloud forecasts produced by the attention-based deep neural networks (SAConvLSTM and CBAMConvLSTM) lead to improved solar forecasts compared to the standard ConvLSTM derived cloud forecasts, and using CBAMConvLSTM cloud forecasts improving over SAConvLSTM. 

An interesting observation can be made regarding the importance of the quality of the cloud forecasts based on the type of solar forecasting network. We can see that for a simple MLP network including cloud predictions always improve the solar forecasts compared to not including clouds. However, for CNN and LSTM networks including cloud forecasts does not always improve the solar forecasts which indicates the importance of the cloud prediction accuracy (i.e., forecasted cloud values being closer to the ground-truth cloud values). 

We further investigate this observation by evaluating the impact of cloud forecasts under different sky conditions inferred through the satellite images:
\begin{enumerate}
    \item Clear sky (i.e., pixel value of $0$) - only considering test samples with clear sky conditions, and
    \item Cloudy (i.e., pixel value $> 0$) - only considering test samples where clouds are present.
\end{enumerate}
We further divide the cloudy condition into two conditions as high altitude clouds and low altitude clouds. For this categorization we consider a pixel value of $50$, if the cloud pixel covering the PV site is greater than $50$, we consider it as a high altitude cloud condition (as we discussed above a brighter colour in the infrared image reflects higher altitude clouds), while a pixel value less than or equal $50$ and greater than $0$ is a low altitude cloud condition. Table \ref{tab:test_samples} shows the percentage of test samples available for the four sky conditions from all test samples under study.

\begin{table}[!ht]
    \centering
    \begin{tabular}{cr}
    \hline
         Sky condition & Percentage of test samples \\ 
    \hline
         Clear sky & 46.57 \% \\
         High altitude clouds & 16.24 \% \\
         Low altitude clouds & 37.19 \% \\
         Cloudy (all clouds) & 53.43 \% \\ 
    \hline
    \end{tabular}
    \caption{Average percentage of clear sky, high altitude clouds, low altitude clouds, and all cloudy conditions present across all test samples of the 50 PV sites under study.}
    \label{tab:test_samples}
\end{table}

\begin{table}[htb]
     \centering
    \begin{tabular}{p{6cm}>{\raggedleft\arraybackslash}p{1cm}>{\raggedleft\arraybackslash}p{1.2cm}>{\raggedleft\arraybackslash}p{1cm}>{\raggedleft\arraybackslash}p{1.2cm}>{\raggedleft\arraybackslash}p{1cm}>{\raggedleft\arraybackslash}p{1.2cm}}
    \hline
   Cloud Input During & \multicolumn{6}{c}{Solar Power Forecasting Network} \\
    Forecasting Stage &   \multicolumn{2}{c}{MLP} & \multicolumn{2}{c}{CNN} & \multicolumn{2}{c}{LSTM} \\
    \cline{2-7}
    & \footnotesize \textit{RMSE Skill Score (\%)} & \footnotesize \textit{Skill \newline Score \newline Difference} & \footnotesize \textit{RMSE Skill Score (\%)} & \footnotesize \textit{Skill \newline Score \newline Difference} & \footnotesize \textit{RMSE Skill Score (\%)} & \footnotesize \textit{Skill \newline Score \newline Difference}  \\
    \hline
    Ground-truth clouds & 51.11 & - & 51.41 & - & 62.47 & - \\
    Forecasted clouds from \newline ConvLSTM & 50.56 & 0.55 & 50.78 & 0.63 & 61.80 & 0.67\\ 
    Forecasted clouds from \newline SAConvLSTM & 50.41 & 0.70 & 50.91 & 0.50 & 61.60 & 0.87 \\ 
    Forecasted clouds from \newline CBAMConvLSTM & 50.56 & \textbf{0.55} & 50.99 & \textbf{0.42} & 61.84 & \textbf{0.63} \\ 
    Persistence clouds & 50.36 & 0.75 & 50.59 & 0.82 & 61.67 & 0.80 \\ 
    No clouds included & 29.27 & 21.84 & 50.23 & 1.18 & 59.87 & 2.60 \\
    \hline
    \end{tabular}
    \caption{Average RMSE Skill Score across 50 rooftop solar PV sites considering only clear sky test samples of a site and the respective skill score difference with regard to using \textit{Ground-truth clouds} showing the best skill score that can be achieved. The lowest skill score difference is marked in boldface.}
    \label{tab:results_clear_sky}
\end{table}

\begin{table}[htb]
     \centering
    \begin{tabular}{p{6cm}>{\raggedleft\arraybackslash}p{1cm}>{\raggedleft\arraybackslash}p{1.2cm}>{\raggedleft\arraybackslash}p{1cm}>{\raggedleft\arraybackslash}p{1.2cm}>{\raggedleft\arraybackslash}p{1cm}>{\raggedleft\arraybackslash}p{1.2cm}}
    \hline
   Cloud Input During  & \multicolumn{6}{c}{Solar Power Forecasting Network} \\
    Forecasting Stage &   \multicolumn{2}{c}{MLP} & \multicolumn{2}{c}{CNN} & \multicolumn{2}{c}{LSTM} \\
    \cline{2-7}
    & \footnotesize \textit{RMSE Skill Score (\%)} & \footnotesize \textit{Skill \newline Score \newline Difference} & \footnotesize \textit{RMSE Skill Score (\%)} & \footnotesize \textit{Skill \newline Score \newline Difference} & \footnotesize \textit{RMSE Skill Score (\%)} & \footnotesize \textit{Skill \newline Score \newline Difference} \\
    \hline
    Ground-truth clouds & 17.56 & - & 13.23 & - & 20.7 & -\\
    Forecasted clouds from \newline ConvLSTM & 9.76 & 7.80 & 6.20 & 7.03 & 11.79 & 8.91\\ 
    Forecasted clouds from \newline SAConvLSTM & 15.73 & 1.83 & 11.64 & 1.59 & 18.09 & 2.61 \\ 
    Forecasted clouds from \newline CBAMConvLSTM & 15.74 & \textbf{1.82} & 12.06 & \textbf{1.17} & 18.46 & \textbf{2.24} \\
    Persistence clouds & 10.38 & 7.18 & 5.07 & 8.16 & 11.04 & 9.66 \\ 
    No clouds included & 3.36 & 14.20 & 11.28 & 1.95 & 16.17 & 4.53\\
    \hline
    \end{tabular}
    \caption{Average RMSE Skill Score across 50 rooftop solar PV sites considering only high altitude cloud condition test samples of a site and the respective skill score difference with regard to using \textit{Ground-truth clouds} showing the best skill score that can be achieved. The lowest skill score difference is marked in boldface.}
    \label{tab:results_high_clouds}
\end{table}

\begin{table}[htb]
    \centering
    \begin{tabular}{p{6cm}>{\raggedleft\arraybackslash}p{1cm}>{\raggedleft\arraybackslash}p{1.2cm}>{\raggedleft\arraybackslash}p{1cm}>{\raggedleft\arraybackslash}p{1.2cm}>{\raggedleft\arraybackslash}p{1cm}>{\raggedleft\arraybackslash}p{1.2cm}}
    \hline
   Cloud Input During & \multicolumn{6}{c}{Solar Power Forecasting Network} \\
    Forecasting Stage &   \multicolumn{2}{c}{MLP} & \multicolumn{2}{c}{CNN} & \multicolumn{2}{c}{LSTM} \\
    \cline{2-7}
    & \footnotesize \textit{RMSE Skill Score (\%)} & \footnotesize \textit{Skill \newline Score \newline Difference} & \footnotesize \textit{RMSE Skill Score (\%)} & \footnotesize \textit{Skill \newline Score \newline Difference} & \footnotesize \textit{RMSE Skill Score (\%)} & \footnotesize \textit{Skill \newline Score \newline Difference} \\
    \hline
    Ground-truth clouds & 22.99 & - & 20.7 & - & 24.42 & - \\
    Forecasted clouds from \newline ConvLSTM & 16.16 & 6.83 & 14.34 & 6.36 & 16.79 & 7.63 \\ 
    Forecasted clouds from \newline SAConvLSTM & 17.28 & 5.71 & 15.22 & 5.48 & 17.90 & 6.52\\ 
    Forecasted clouds from \newline CBAMConvLSTM & 17.69 & \textbf{5.30} & 15.75 & 4.95 & 18.29 & 6.13 \\ 
    Persistence clouds & 12.81 & 10.18  & 10.64 & 10.06 & 13.72 & 10.70\\ 
    No clouds included & 12.12 & 10.87 & 18.05 & \textbf{2.65} & 20.00 & \textbf{4.42} \\
    \hline
    \end{tabular}
    \caption{Average RMSE Skill Score across 50 rooftop solar PV sites considering only low altitude cloud condition test samples of a site and the respective skill score difference with regard to using \textit{Ground-truth clouds} showing the best skill score that can be achieved. The lowest skill score difference is marked in boldface.}
    \label{tab:results_low_clouds}
\end{table}

\begin{table}[htb]
     \centering
    \begin{tabular}{p{6cm}>{\raggedleft\arraybackslash}p{1cm}>{\raggedleft\arraybackslash}p{1.2cm}>{\raggedleft\arraybackslash}p{1cm}>{\raggedleft\arraybackslash}p{1.2cm}>{\raggedleft\arraybackslash}p{1cm}>{\raggedleft\arraybackslash}p{1.2cm}}
    \hline
   Cloud Input During & \multicolumn{6}{c}{Solar Power Forecasting Network} \\
    Forecasting Stage &   \multicolumn{2}{c}{MLP} & \multicolumn{2}{c}{CNN} & \multicolumn{2}{c}{LSTM} \\
    \cline{2-7}
    & \footnotesize \textit{RMSE Skill Score (\%)} & \footnotesize \textit{Skill \newline Score \newline Difference} & \footnotesize \textit{RMSE Skill Score (\%)} & \footnotesize \textit{Skill \newline Score \newline Difference} & \footnotesize \textit{RMSE Skill Score (\%)} & \footnotesize \textit{Skill \newline Score \newline Difference} \\
    \hline
    Ground-truth clouds & 21.5 & - & 18.64 & - & 23.39 & - \\
    Forecasted clouds from \newline ConvLSTM & 14.39 & 7.11 & 12.08 & 6.56 & 15.4 & 7.99 \\ 
    Forecasted clouds from \newline SAConvLSTM & 16.88 & 4.62 & 14.25 & 4.39 & 17.96 & 5.43\\ 
    Forecasted clouds from \newline CBAMConvLSTM & 17.17 & \textbf{4.33} & 14.74 & 3.90 & 18.35 & 5.04\\
    Persistence clouds & 12.15 & 9.35 & 9.10 & 9.54 & 12.97 & 10.42\\ 
    No clouds included & 9.67 & 11.83 & 16.17 & \textbf{2.47} & 18.93 & \textbf{4.46}\\
    \hline
    \end{tabular}
    \caption{Average RMSE Skill Score across 50 rooftop solar PV sites considering only sky conditions where clouds are present in test samples of a site and the respective skill score difference with regard to using \textit{Ground-truth clouds} showing the best skill score that can be achieved. The lowest skill score difference is marked in boldface.}
    \label{tab:results_all_clouds}
\end{table}

Tables \ref{tab:results_clear_sky}, \ref{tab:results_high_clouds}, \ref{tab:results_low_clouds}, \ref{tab:results_all_clouds} shows the RMSE skill scores for clear sky, high altitude clouds, low altitude clouds and cloudy conditions respectively. Under clear sky conditions, forecasting clouds using a deep neural network or persistence clouds does not make a difference as expected. Moreover, a higher skill score can be achieved as there are no clouds that impact solar generation. LSTM and CNN trained without including clouds achieve a skill score closer to the scenario of including ground-truth clouds. This indicates the ability of CNN and LSTM networks to capture temporal patterns in the data and forecast the solar generation under clear sky conditions learning only through the past hour's power generation patterns. On the contrary, the MLP network can achieve a higher skill score if trained using cloud inputs. This could be due to the inability of the MLP network to capture temporal patterns in the solar generation time series and therefore, challenging to learn only using past hour's solar generation information.  

Under high altitude cloud conditions (Table \ref{tab:results_high_clouds}), solar forecast from all three networks are improved by including ground-truth clouds which indicate that if the cloud predictions are closer to the ground-truth cloud values the solar forecasts can be improved. For all three networks, using CBAMConvLSTM cloud forecasts as an input result in a skill score closer to the ground-truth cloud condition followed by the SAConvLSTM. However, including cloud forecasts from ConvLSTM or persistence clouds deteriorates the forecasts compared to training without any clouds. These observations emphasise to the importance of having accurate cloud forecasts when forecasting solar generation. 

Under low altitude cloud conditions (Table \ref{tab:results_low_clouds}), similar to high altitude conditions, solar forecasts from all three networks are improved when ground-truth cloud are included. However, including cloud forecasts have increased the skill score closer to ground-truth cloud's skill score only for the MLP network. For LSTM and CNN networks using cloud predictions from any deep neural network or persistence clouds deteriorate the solar forecasts. This could also be because when clouds affecting the solar generation are at lower altitudes, the impact on the solar generation is high (i.e., the generation is lower) and therefore, when the cloud input is deviated from the ground-truth value, CNN and LSTM networks are unable predict the low solar generation accurately. Nonetheless, the LSTM network using cloud predictions from attention-based methods, CBAMConvLSTM and SAConvLSTM achieves higher skill scores in all cloud input scenarios compared to the MLP network. 

Table \ref{tab:results_all_clouds} summarizes the results considering all cloudy test samples. The results show a similar observations as the low altitude cloud test samples where including ground-truth clouds can improve the solar forecasts but using cloud forecasts can deteriorate the solar forecasts for CNN and LSTM networks. These results are consistent with low altitude clouds as majority of the cloudy test samples consist of low altitude cloud conditions as seen in Table \ref{tab:test_samples}. 
\revised{
It is important to note that ground-truth cloud observations are unavailable at the time of prediction in real-world operational settings. In practice, backtesting forecast models across historical  periods allows to systematically understand the performance of each method under different conditions and inputs. In such scenarios, where systematic variations are observed across models under different conditions, an ensemble framework where the outputs of multiple models utilising different inputs are collectively used in generating the final forecast can be considered as a practical solution.
}

\begin{figure}[!htb]
    \centering
    \includegraphics[width=0.9\textwidth]{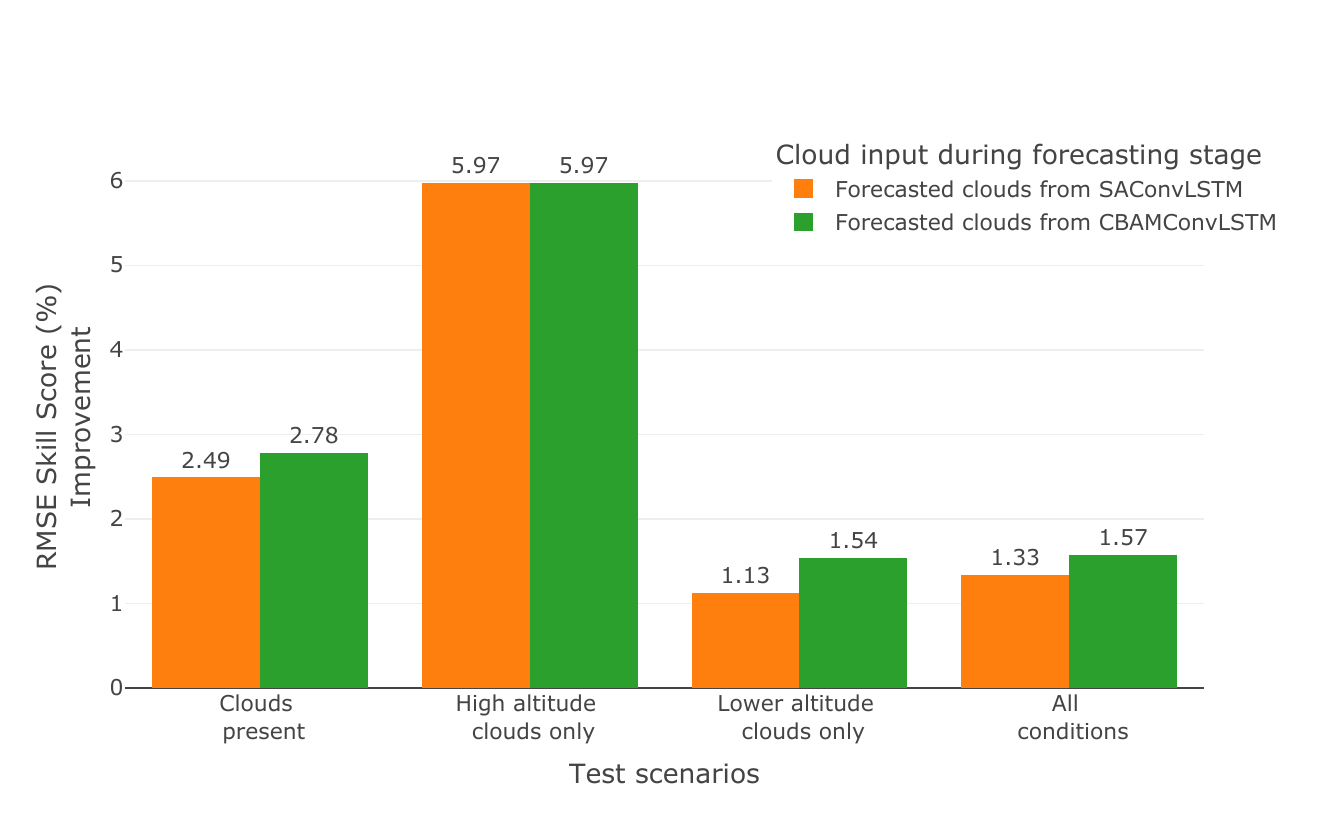}
    \caption{Improvement in the RMSE skill score of the MLP network under different test scenarios when using cloud forecasts from attention-based methods - SAConvLSTM and CBAMConvLSTM compared to using cloud forecasts from ConvLSTM.}
    \label{fig:mlp_results}
\end{figure}

\begin{figure}[!htb]
    \centering
    \includegraphics[width=0.9\textwidth]{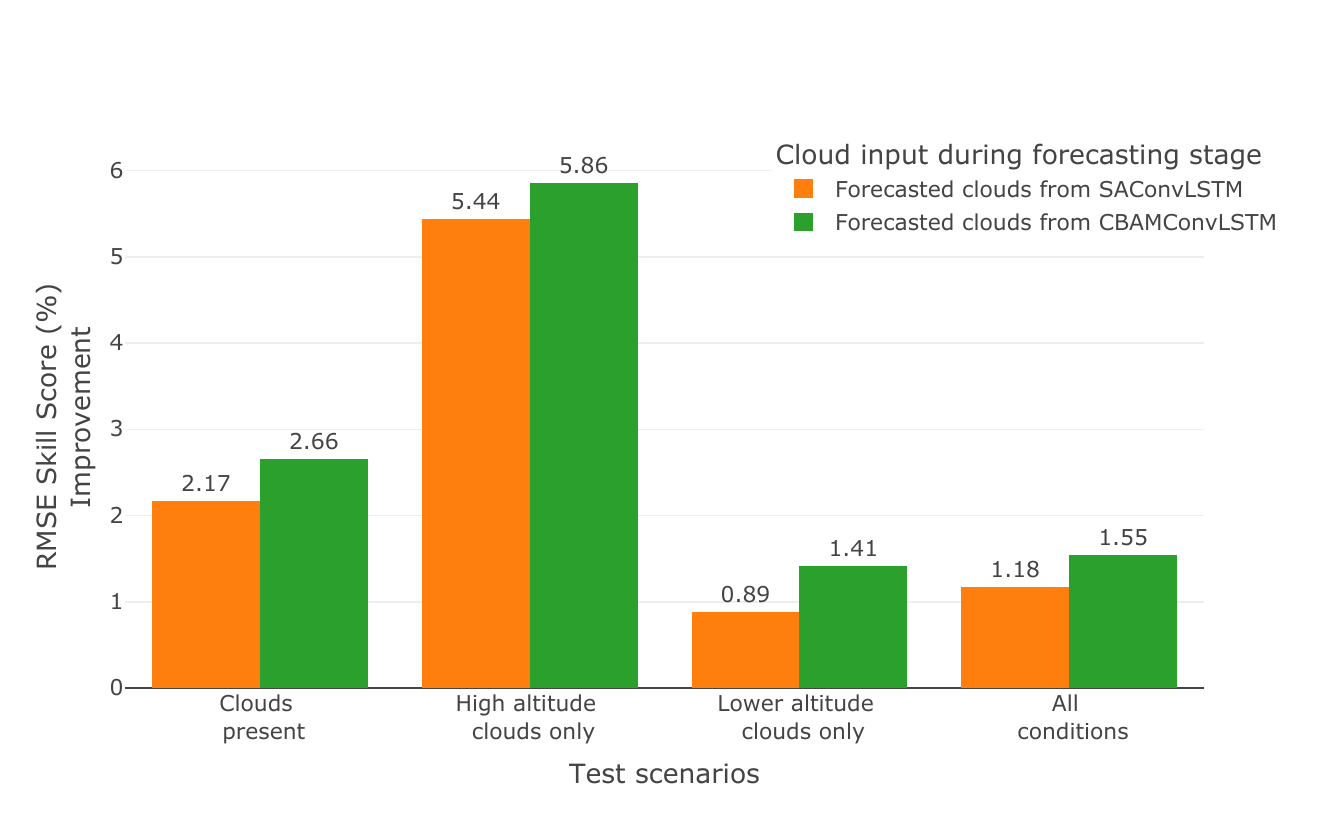}
    \caption{Improvement in the RMSE skill score of the CNN network under different test scenarios when using cloud forecasts from attention-based methods - SAConvLSTM and CBAMConvLSTM compared to using cloud forecasts from ConvLSTM.}
    \label{fig:cnn_results}
\end{figure}

\begin{figure}[!htb]
    \centering
    \includegraphics[width=0.9\textwidth]{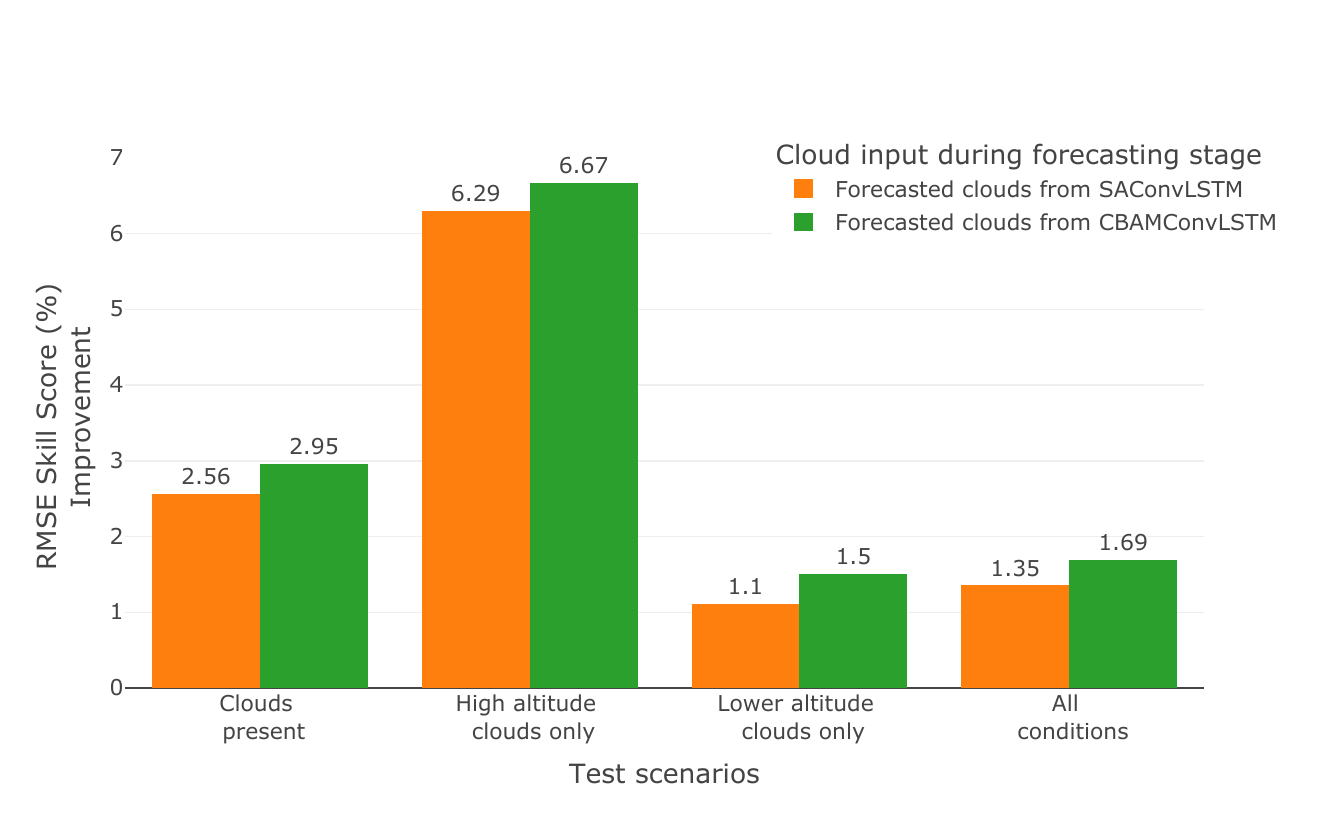}
    \caption{Improvement in the RMSE skill score of the LSTM network under different test scenarios when using cloud forecasts from attention-based methods - SAConvLSTM and CBAMConvLSTM compared to using cloud forecasts from ConvLSTM.}
    \label{fig:lstm_results}
\end{figure}

\subsection{Comparison of attention-based and non-attention cloud forecasting}

Figures \ref{fig:mlp_results}, \ref{fig:cnn_results}, \ref{fig:lstm_results} show the RMSE skill score improvements when using cloud forecasts from CBAMConvLSTM, SAConvLSTM compared to ConvLSTM. We can see that using cloud forecasts from CBAMConvLSTM and SAConvLSTM have improved the skill scores compared to using cloud forecasts from ConvLSTM under all sky conditions. In test samples where clouds are present, CBAMConvLSTM increases the RMSE skill score by 2.78, 2.66, 2.95 for MLP, CNN, LSTM networks respectively in comparison to ConvLSTM. We can observe the best improvement with attention-based networks in high altitude cloud test samples where CBAMConvLSTM cloud inputs result in a skill score increase of 5.97 (MLP), 5.86 (CNN), 6.67 (LSTM) and SAConvLSTM result in a skill score increase of 5.97 (MLP), 5.44 (CNN), 6.29 (LSTM). Under low altitude cloud test samples the skill score increase is comparatively lower where CBAMConvLSTM cloud inputs result in a skill score increase of 1.54 (MLP), 1.41 (CNN), 1.50 (LSTM) and SAConvLSTM result in a skill score increase of 1.13 (MLP), 0.89 (CNN), 1.10 (LSTM). It is noteworthy, that we consider high altitude cloud scenarios in this work when pixel values greater than 50. These values appear in colors closer to white in the infrared satellite image. Therefore, it is possible that the attention-based networks have been successful predicting pixel values closer to shades of white in the satellite image compared to other areas contributing to the skill score improvements in high altitude cloud scenarios.

\begin{figure}[!ht]
     \centering
     \begin{subfigure}[]{\textwidth}
         \centering
         \includegraphics[width=\textwidth]{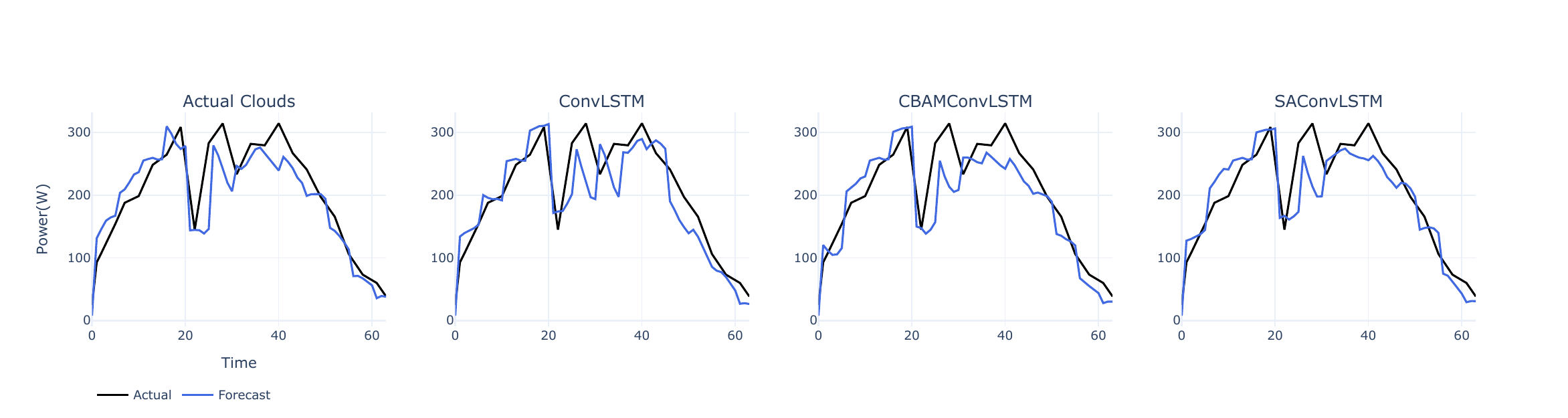}
         \caption{Solar PV site 1.}
         \label{fig:forecast1}
     \end{subfigure}
      \begin{subfigure}[]{\textwidth}
         \centering
         \includegraphics[width=\textwidth]{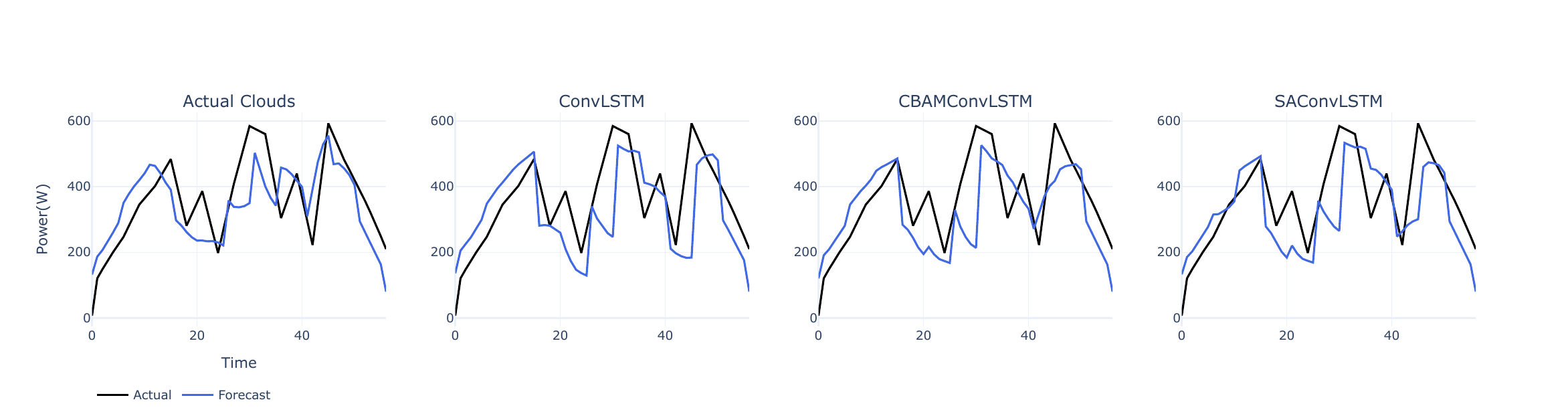}
         \caption{Solar PV site 2.}
         \label{fig:forecast2}
     \end{subfigure}
        \caption{\revised{A comparison of actual solar generation from two solar sites with forecasts generated by the LSTM solar forecasting network using actual cloud observations and cloud forecasts from ConvLSTM, CBAMConvLSTM and SAConvLSTM.}}
        \label{fig:forecast_visualisation}
\end{figure}

\revised{
Figure \ref{fig:forecast_visualisation} presents a comparison of actual solar generation at two sites against solar forecasts produced by the LSTM network, using both actual cloud observations and cloud forecasts derived from ConvLSTM, CBAMConvLSTM, and SAConvLSTM.
As shown in Figure \ref{fig:forecast1}, CBAMConvLSTM and SAConvLSTM more accurately capture the rapid fluctuations in solar generation (e.g., time indices 20–50) compared to ConvLSTM-based cloud forecasts. Similarly, Figure \ref{fig:forecast2} demonstrates that the fluctuations around time index 40 are more accurately captured when using cloud forecasts from CBAMConvLSTM and SAConvLSTM, whereas ConvLSTM consistently under-forecasts during this period.
} 

\subsection{\revised{Limitations and Future Directions}}

In this work, as discussed above in Section \ref{sec:framework}, we focused on a two-step process to forecast solar generation using cloud predictions. Therefore, we trained attention-based networks to predict a sequence of infrared satellite images given a past sequence of the images. Subsequently, a pixel in the predicted images corresponding to the latitude and longitude of a PV site was extracted as the clouds covering the PV site to train the solar forecasting methods. These extracted pixels represents the clouds vertically above the PV site as we discussed before. However, clouds covering the sunlight may not be vertically above the site depending on the angle of sunlight reaching the ground \cite{Si2021PhotovoltaicPosition}. As the satellite images covered a 2 km spatial resolution, a pixel in the image covered a relatively large area and therefore, taking the pixel matching the latitude and longitude was reasonable. However, this approach may be extended by finding the exact cloud pixel covering the PV site using information such as solar azimuth, zenith, and cloud heights. 

We further divided the cloud conditions as high and low altitude clouds based on the pixel values of the satellite image. However, this categorization may not precisely reflect high and low altitude clouds as presented in some works such as \cite{Barbieri2017VeryReview} as the cloud altitude is not explicitly calculated in our work.

The work presented in this paper focuses on infrared satellite images \revised{(band 13)} that can be used to infer information on cloud altitudes. As a natural extension, future studies could explore \revised{other satellite data sources}, such as visible imagery, which captures cloud thickness information, or \revised{\textit{Himawari-based Shortwave Radiation} products that more directly represent the solar irradiance reaching earth surface. Additionally, given that brightness temperature observations in infrared images may carry residual diurnal patterns beyond cloud effects, investigating pre-processing strategies to better isolate cloud-related information represents another worthwhile direction.}

\revised{
Furthermore, this work focuses on short-term solar generation forecasting via an indirect two-stage approach, in which cloud forecasting serves as an interpretable intermediate step, enabling the impact of attention-based mechanisms on cloud prediction to be systematically assessed. As discussed in Section \ref{sec:relatedwork}, direct forecasting methods also exist that consolidate the cloud and solar forecasting stages into a single end-to-end pipeline. A rigorous comparison with such direct attention-based methods for solar generation forecasting, together with a systematic analysis of computational costs across approaches, represents a valuable and natural extension of this work.
}

\revised{Recently, novel deep neural network architectures}, such as PredRNN \cite{Wang2022PredRNN:Learning} have been proposed for spatio-temporal applications to overcome the existing limitations of standard ConvLSTM. As future work, it would be interesting to investigate the adaption of such networks and their ability to forecast cloud movement using satellite images. 

\section{Conclusions}
\label{sec:conclusion}

The movement of clouds is the primary driver of short-term fluctuations of solar generation. In this work, we studied the \revised{impact of using cloud forecasts generated from attention and non-attention based deep neural networks on the down-stream task of short-term solar generation forecasting.} \revised{To empirically investigate how attention-based  improvements in cloud forecasting impact solar forecasting}, we proposed an attention-based deep neural network (CBAMConvLSTM) \revised{which extends the ConvLSTM network widely used for spatio-temporal prediction.} In addition, we adapted another attention-based deep neural network (SAConvLSTM) originally developed for video prediction. Both attention-based networks were compared against the non-attention based \revised{baseline ConvLSTM}. The impact of cloud forecasts when forecasting solar generation was evaluated for 50 small-scale rooftop solar PV sites in Australia, using three state-of-the-art neural networks (MLP, CNN, LSTM) trained with and without cloud values as an input to the networks. The quality of cloud predictions from CBAMConvLSTM, SAConvLSTM and ConvLSTM were evaluated against using ground-truth and persistence cloud values as inputs to the solar forecasting networks. 

Our results demonstrated that \revised{using cloud forecasts from attention-based networks achieved a higher RMSE solar forecast skill score compared to using cloud forecasts from non-attention-based ConvLSTM and using LSTM with cloud predictions from CBAMConvLSTM as input to the LSTM network achieves the best solar power forecasts.} When analyzed across different sky conditions, it was observed that in clear sky conditions, the inclusion of cloud forecasts (ground-truth or predicted) does not make much of a difference in the solar forecast accuracy. However, for intervals having cloudy skies, including cloud forecasts as an input to solar forecasting methods resulted in higher solar forecast skill score in most scenarios.  In particular, using cloud predictions from the proposed CBAMConvLSTM improved the RMSE skill score by 5.86\%, 5.97\%, 6.67\% for MLP, CNN and LSTM networks respectively, while SAConvLSTM improved the RMSE skill score 5.44\%, 5.97\%, 6.29\% under high altitude cloud scenarios (i.e., pixel values greater than 50 in a gray-scale infrared image). However, the improvements under lower altitude conditions (i.e., pixel values less than 50) were minimal. 

\section*{Acknowledgement}
The authors are grateful to Solar Analytics for providing anonymised solar power generation data to conduct this research. This work was supported by the Melbourne Research Scholarship awarded to the first author. SH and JDH acknowledge Australian Research Council grant DP220101035. The research was undertaken using the LIEF HPC-GPGPU Facility hosted at the University of Melbourne (this facility was established with the assistance of LIEF Grant LE170100200). 

\appendix
\include{Appendix}



\bibliographystyle{elsarticle-num-names} 
\bibliography{references.bib}

\end{document}

%% file: Appendix.tex
\section{Average MAE Skill Scores}
\label{appendixA}

\begin{figure}[!htb]
    \centering
    \includegraphics[width=0.8\textwidth]{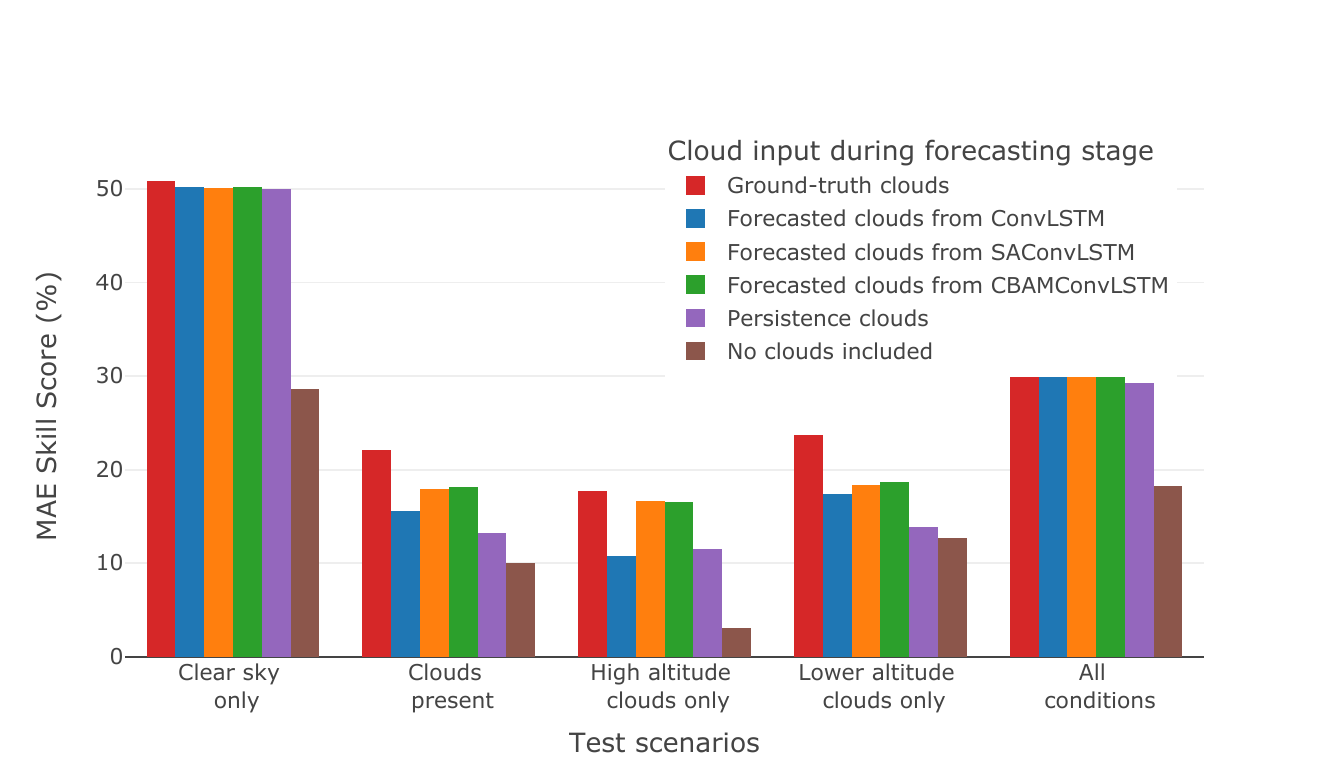}
    \caption{Average MAE Skill Score across 50 PV sites from the MLP network under different test scenarios and cloud inputs.}
    \label{fig:mlp_mae}
\end{figure}

\begin{figure}[htb]
    \centering
    \includegraphics[width=0.8\textwidth]{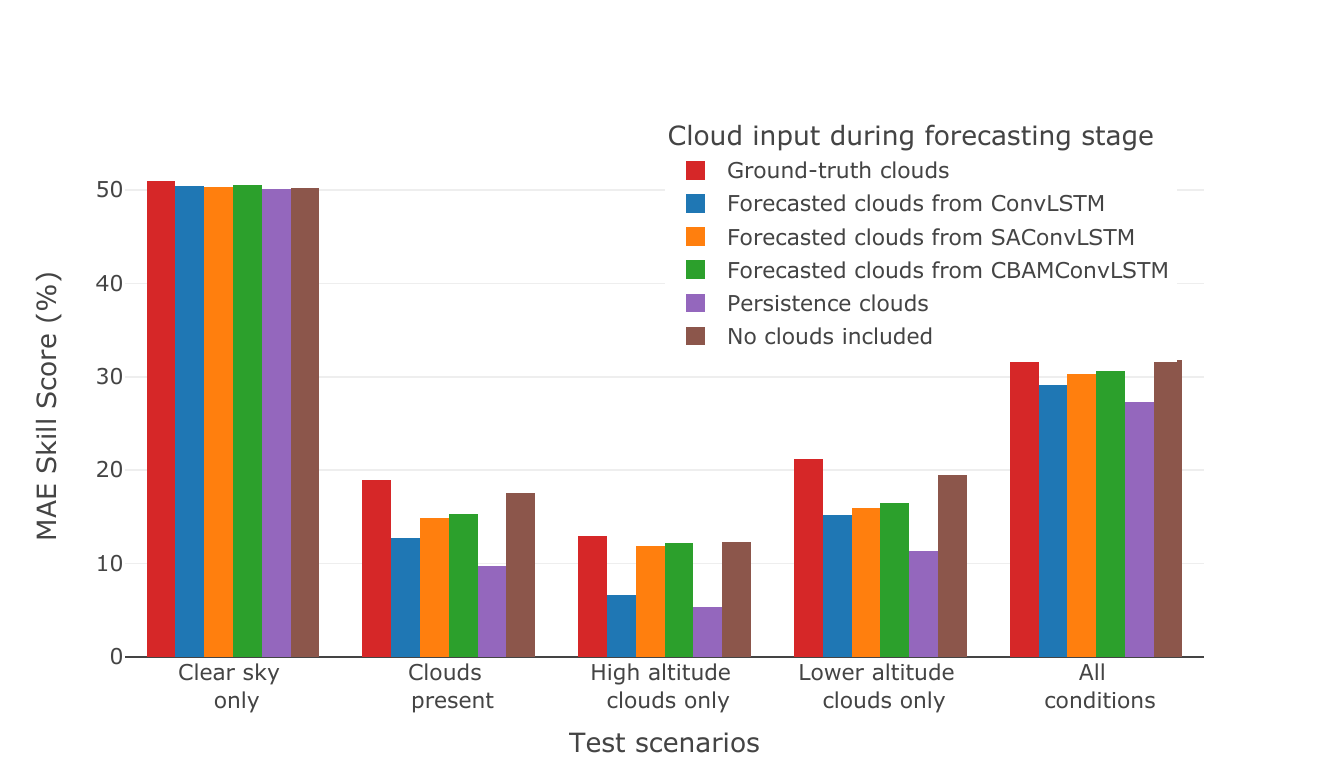}
    \caption{Average MAE Skill Score across 50 PV sites from the CNN network under different test scenarios and cloud inputs.}
    \label{fig:cnn_mae}
\end{figure}

\begin{figure}[htb]
    \centering
    \includegraphics[width=0.8\textwidth]{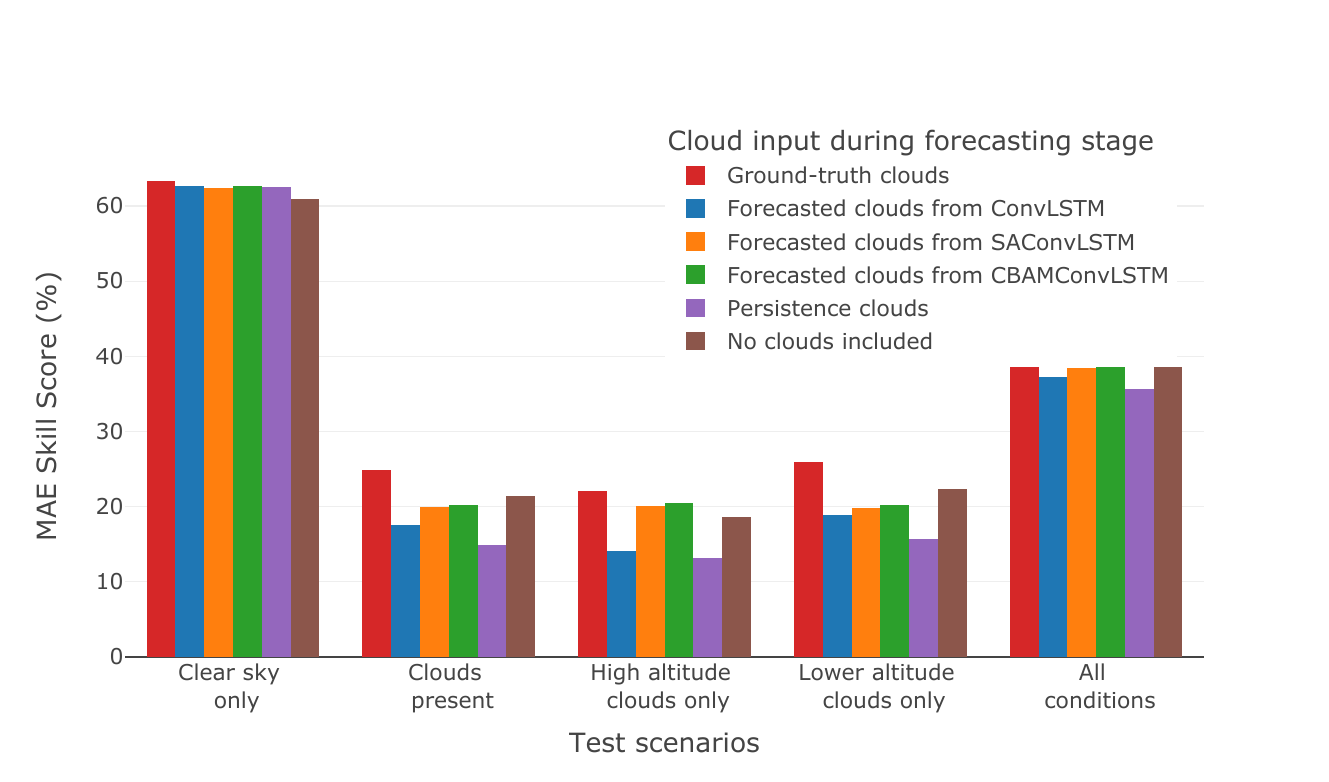}
    \caption{Average MAE Skill Score across 50 PV sites from the LSTM network under different test scenarios and cloud inputs.}
    \label{fig:lstm_mae}
\end{figure}

Table \ref{tab:results_mae} shows the average RMSE skill score considering all test samples under all sky conditions. Figures \ref{fig:mlp_mae}, \ref{fig:lstm_mae}, \ref{fig:cnn_mae} shows the average MAE skill scores for MLP, CNN and LSTM networks under test scenarios of different sky conditions studied in this work. 

\begin{table}[htb]
    \centering
    \begin{tabular}{p{5cm}>{\raggedleft\arraybackslash}p{1.5cm}>{\raggedleft\arraybackslash}p{1.2cm}>{\raggedleft\arraybackslash}p{1.5cm}>{\raggedleft\arraybackslash}p{1.2cm}>{\raggedleft\arraybackslash}p{1.5cm}>{\raggedleft\arraybackslash}p{1.2cm}}
    \hline
   Cloud Input During  & \multicolumn{6}{c}{Solar Power Forecasting Network} \\
    Forecasting Stage &   \multicolumn{2}{c}{MLP} & \multicolumn{2}{c}{CNN} & \multicolumn{2}{c}{LSTM} \\
     \cline{2-7}
    & \footnotesize \textit{MAE Skill Score (\%)} & \footnotesize \textit{Skill \newline Score \newline Difference} & \footnotesize \textit{MAE Skill Score (\%)} & \footnotesize \textit{Skill \newline Score \newline Difference} & \footnotesize \textit{MAE Skill Score (\%)} & \footnotesize \textit{Skill \newline Score \newline Difference} \\
    \hline
    Ground-truth clouds & 34.58 & - & 32.84 & - & 41.64 & -\\
    Forecasted clouds from \newline ConvLSTM & 30.65 & 3.93 & 29.17 & 3.67 & 37.20 & 4.44\\ 
    Forecasted clouds from \newline SAConvLSTM & 31.90 & 2.68 & 30.28 & 2.56 & 38.45 & 3.19\\ 
    Forecasted clouds from \newline CBAMConvLSTM & 32.10 & \textbf{2.48} & 30.62 & 2.22 & 38.76 & \textbf{2.88}\\ 
    Persistence clouds & 29.23 & 5.35 & 27.29 & 5.55 & 35.66 & 5.59 \\ 
    No clouds included & 18.25 & 16.33 & 31.76 & \textbf{1.08} & 38.57 & 3.07\\
    \hline
    \end{tabular}
    \caption{Average MAE Skill Score across 50 rooftop solar PV sites considering all test samples of a site and the respective skill score difference with regard to using \textit{Ground-truth clouds} showing the best skill score that can be achieved. The lowest skill score difference is marked in boldface.}
    \label{tab:results_mae}
\end{table}

%% file: references.bib
@article{Zhao20193D-CNN-basedPrediction,
    title = {{3D-CNN-based feature extraction of ground-based cloud images for direct normal irradiance prediction}},
    year = {2019},
    journal = {Solar Energy},
    author = {Zhao, Xin and Wei, Haikun and Wang, Hai and Zhu, Tingting and Zhang, Kanjian},
    number = {February},
    pages = {510--518},
    volume = {181},
    publisher = {Elsevier},
    url = {https://doi.org/10.1016/j.solener.2019.01.096},
    doi = {10.1016/j.solener.2019.01.096},
    issn = {0038092X}
}

@article{Castangia2021AForecasting,
    title = {{A compound of feature selection techniques to improve solar radiation forecasting}},
    year = {2021},
    journal = {Expert Systems with Applications},
    author = {Castangia, Marco and Aliberti, Alessandro and Bottaccioli, Lorenzo and Macii, Enrico and Patti, Edoardo},
    month = {9},
    pages = {114979},
    volume = {178},
    publisher = {Pergamon},
    doi = {10.1016/j.eswa.2021.114979},
    issn = {09574174}
}

@article{Lai2021AForecasting,
    title = {{A deep learning based hybrid method for hourly solar radiation forecasting}},
    year = {2021},
    journal = {Expert Systems with Applications},
    author = {Lai, Chun Sing and Zhong, Cankun and Pan, Keda and Ng, Wing W.Y. and Lai, Loi Lei},
    month = {9},
    pages = {114941},
    volume = {177},
    publisher = {Pergamon},
    doi = {10.1016/j.eswa.2021.114941},
    issn = {09574174}
}

@inproceedings{Bansal2022ALearning,
    title = {{A Moment in the Sun: Solar Nowcasting from Multispectral Satellite Data using Self-Supervised Learning}},
    year = {2022},
    booktitle = {Proceedings of the Thirteenth ACM International Conference on Future Energy Systems},
    author = {Bansal, Akansha Singh and Bansal, Trapit and Irwin, David},
    pages = {251--262},
    url = {http://arxiv.org/abs/2112.13974},
    arxivId = {2112.13974}
}

@article{Erdener2022AForecasting,
    title = {{A review of behind-the-meter solar forecasting}},
    year = {2022},
    journal = {Renewable and Sustainable Energy Reviews},
    author = {Erdener, Burcin Cakir and Feng, Cong and Doubleday, Kate and Florita, Anthony and Hodge, Bri Mathias},
    number = {November 2021},
    pages = {112224},
    volume = {160},
    publisher = {Elsevier Ltd},
    url = {https://doi.org/10.1016/j.rser.2022.112224},
    doi = {10.1016/j.rser.2022.112224},
    issn = {18790690}
}

@article{Liu2017AApplications,
    title = {{A survey of deep neural network architectures and their applications}},
    year = {2017},
    journal = {Neurocomputing},
    author = {Liu, Weibo and Wang, Zidong and Liu, Xiaohui and Zeng, Nianyin and Liu, Yurong and Alsaadi, Fuad E.},
    month = {4},
    pages = {11--26},
    volume = {234},
    publisher = {Elsevier},
    doi = {10.1016/J.NEUCOM.2016.12.038},
    issn = {0925-2312}
}

@article{Chaudhari2021AnModels,
    title = {{An Attentive Survey of Attention Models}},
    year = {2021},
    journal = {ACM Transactions on Intelligent Systems and Technology},
    author = {Chaudhari, Sneha and Mithal, Varun and Polatkan, Gungor and Ramanath, Rohan},
    number = {5},
    pages = {53},
    volume = {12},
    url = {https://doi.org/10.1145/3465055},
    doi = {10.1145/3465055},
    issn = {21576912},
    arxivId = {1904.02874}
}

@inproceedings{VaswaniAshishandShazeerNoamandParmarNikiandUszkoreitJakobandJonesLlionandGomezAidanNandKaiserLukaszandPolosukhin2017AttentionNeed,
    title = {{Attention Is All You Need}},
    year = {2017},
    booktitle = {Advances in Neural Information Processing Systems},
    author = {Vaswani, Ashish and Shazeer, Noam and Parmar, Niki and Uszkoreit, Jakob and Jones, Llion and Gomez, Aidan N and Kaiser, {\{}{\textbackslash}L{\}}ukasz and Polosukhin, Illia},
    volume = {30},
    doi = {10.1109/2943.974352},
    issn = {10772618}
}

@article{Guo2022AttentionSurvey,
    title = {{Attention mechanisms in computer vision: A survey}},
    year = {2022},
    journal = {Computational Visual Media},
    author = {Guo, Meng Hao and Xu, Tian Xing and Liu, Jiang Jiang and Liu, Zheng Ning and Jiang, Peng Tao and Mu, Tai Jiang and Zhang, Song Hai and Martin, Ralph R. and Cheng, Ming Ming and Hu, Shi Min},
    number = {3},
    month = {9},
    pages = {331--368},
    volume = {8},
    publisher = {Tsinghua University},
    doi = {10.1007/s41095-022-0271-y},
    issn = {20960662},
    arxivId = {2111.07624}
}

@article{Paletta2021BenchmarkingAnalysis,
    title = {{Benchmarking of deep learning irradiance forecasting models from sky images – An in-depth analysis}},
    year = {2021},
    journal = {Solar Energy},
    author = {Paletta, Quentin and Arbod, Guillaume and Lasenby, Joan},
    number = {June},
    pages = {855--867},
    volume = {224},
    publisher = {Elsevier Ltd},
    url = {https://doi.org/10.1016/j.solener.2021.05.056},
    doi = {10.1016/j.solener.2021.05.056},
    issn = {0038092X},
    arxivId = {2102.00721}
}

@inproceedings{Woo2018CBAM:Module,
    title = {{CBAM: Convolutional block attention module}},
    year = {2018},
    booktitle = {Proceedings of the European conference on computer vision (ECCV)},
    author = {Woo, Sanghyun and Park, Jongchan and Lee, Joon Young and Kweon, In So},
    pages = {3--19},
    isbn = {9783030012335},
    doi = {10.1007/978-3-030-01234-2{\_}1},
    issn = {16113349},
    arxivId = {1807.06521}
}

@article{Dolara2015ComparisonPrediction,
    title = {{Comparison of different physical models for PV power output prediction}},
    year = {2015},
    journal = {Solar Energy},
    author = {Dolara, Alberto and Leva, Sonia and Manzolini, Giampaolo},
    doi = {10.1016/j.solener.2015.06.017},
    issn = {0038092X}
}

@inproceedings{Shi2015ConvolutionalNowcasting,
    title = {{Convolutional LSTM network: A machine learning approach for precipitation nowcasting}},
    year = {2015},
    booktitle = {Advances in Neural Information Processing Systems},
    author = {Shi, Xingjian and Chen, Zhourong and Wang, Hao and Yeung, Dit Yan and Wong, Wai Kin and Woo, Wang Chun},
    pages = {802--810},
    volume = {2015-Janua},
    issn = {10495258},
    arxivId = {1506.04214}
}

@article{Qu2021Day-aheadPattern,
    title = {{Day-ahead hourly photovoltaic power forecasting using attention-based CNN-LSTM neural network embedded with multiple relevant and target variables prediction pattern}},
    year = {2021},
    journal = {Energy},
    author = {Qu, Jiaqi and Qian, Zheng and Pei, Yan},
    pages = {120996},
    volume = {232},
    publisher = {Elsevier Ltd},
    url = {https://doi.org/10.1016/j.energy.2021.120996},
    doi = {10.1016/j.energy.2021.120996},
    issn = {03605442}
}

@article{Ajith2021DeepData,
    title = {{Deep learning based solar radiation micro forecast by fusion of infrared cloud images and radiation data}},
    year = {2021},
    journal = {Applied Energy},
    author = {Ajith, Meenu and Mart{\'{i}}nez-Ram{\'{o}}n, Manel},
    month = {7},
    volume = {294},
    publisher = {Elsevier Ltd},
    doi = {10.1016/J.APENERGY.2021.117014},
    issn = {03062619}
}

@article{Mellit2021DeepForecasting,
    title = {{Deep learning neural networks for short-term photovoltaic power forecasting}},
    year = {2021},
    journal = {Renewable Energy},
    author = {Mellit, A. and Pavan, A. Massi and Lughi, V.},
    pages = {276--288},
    volume = {172},
    publisher = {Elsevier Ltd},
    url = {https://doi.org/10.1016/j.renene.2021.02.166},
    doi = {10.1016/j.renene.2021.02.166},
    issn = {18790682}
}

@article{Zhang2018DeepNowcasting,
    title = {{Deep photovoltaic nowcasting}},
    year = {2018},
    journal = {Solar Energy},
    author = {Zhang, Jinsong and Verschae, Rodrigo and Nobuhara, Shohei and Lalonde, Jean François},
    number = {October},
    pages = {267--276},
    volume = {176},
    publisher = {Elsevier},
    url = {https://doi.org/10.1016/j.solener.2018.10.024},
    doi = {10.1016/j.solener.2018.10.024},
    issn = {0038092X},
    arxivId = {1810.06327}
}

@inproceedings{Huang2017DenselyNetworks,
    title = {{Densely connected convolutional networks}},
    year = {2017},
    booktitle = {Proceedings - 30th IEEE Conference on Computer Vision and Pattern Recognition, CVPR 2017},
    author = {Huang, Gao and Liu, Zhuang and Van Der Maaten, Laurens and Weinberger, Kilian Q.},
    pages = {2261--2269},
    volume = {2017-Janua},
    isbn = {9781538604571},
    doi = {10.1109/CVPR.2017.243},
    arxivId = {1608.06993}
}

@article{Mayer2021ExtensiveForecasting,
    title = {{Extensive comparison of physical models for photovoltaic power forecasting}},
    year = {2021},
    journal = {Applied Energy},
    author = {Mayer, Martin János and Gr{\'{o}}f, Gyula},
    number = {November 2020},
    volume = {283},
    doi = {10.1016/j.apenergy.2020.116239},
    issn = {03062619}
}

@article{Yu2020ForecastingMemory,
    title = {{Forecasting Day-Ahead Hourly Photovoltaic Power Generation Using Convolutional Self-Attention Based Long Short-Term Memory}},
    year = {2020},
    journal = {Energies},
    author = {Yu, Dukhwan and Choi, Wonik and Kim, Myoungsoo and Liu, Ling},
    number = {15},
    month = {8},
    pages = {4017},
    volume = {13},
    publisher = {Multidisciplinary Digital Publishing Institute},
    url = {https://www.mdpi.com/1996-1073/13/15/4017/htm https://www.mdpi.com/1996-1073/13/15/4017},
    doi = {10.3390/EN13154017},
    issn = {1996-1073}
}

@article{Yang2018HistoryMining,
    title = {{History and trends in solar irradiance and PV power forecasting: A preliminary assessment and review using text mining}},
    year = {2018},
    journal = {Solar Energy},
    author = {Yang, Dazhi and Kleissl, Jan and Gueymard, Christian A. and Pedro, Hugo T.C. and Coimbra, Carlos F.M.},
    number = {October 2017},
    pages = {60--101},
    volume = {168},
    publisher = {Elsevier},
    url = {https://doi.org/10.1016/j.solener.2017.11.023},
    doi = {10.1016/j.solener.2017.11.023},
    issn = {0038092X}
}

@article{Kong2020HybridForecasting,
    title = {{Hybrid approaches based on deep whole-sky-image learning to photovoltaic generation forecasting}},
    year = {2020},
    journal = {Applied Energy},
    author = {Kong, Weicong and Jia, Youwei and Dong, Zhao Yang and Meng, Ke and Chai, Songjian},
    number = {August},
    pages = {115875},
    volume = {280},
    publisher = {Elsevier Ltd},
    url = {https://doi.org/10.1016/j.apenergy.2020.115875},
    doi = {10.1016/j.apenergy.2020.115875},
    issn = {03062619}
}

@article{Wang2004ImageSimilarity,
    title = {{Image quality assessment: From error visibility to structural similarity}},
    year = {2004},
    journal = {IEEE Transactions on Image Processing},
    author = {Wang, Zhou and Bovik, Alan Conrad and Sheikh, Hamid Rahim and Simoncelli, Eero P.},
    number = {4},
    month = {4},
    pages = {600--612},
    volume = {13},
    doi = {10.1109/TIP.2003.819861},
    issn = {10577149},
    pmid = {15376593}
}

@article{Svozil1997IntroductionNetworks,
    title = {{Introduction to multi-layer feed-forward neural networks}},
    year = {1997},
    journal = {Chemometrics and Intelligent Laboratory Systems},
    author = {Svozil, Daniel and Kvasni{\v{c}}ka, Vladimír and Posp{\'{i}}chal, Jiří},
    number = {1},
    month = {11},
    pages = {43--62},
    volume = {39},
    publisher = {Elsevier},
    doi = {10.1016/S0169-7439(97)00061-0},
    issn = {0169-7439}
}

@article{Kumari2021LongForecasting,
    title = {{Long short term memory–convolutional neural network based deep hybrid approach for solar irradiance forecasting}},
    year = {2021},
    journal = {Applied Energy},
    author = {Kumari, Pratima and Toshniwal, Durga},
    number = {January},
    pages = {117061},
    volume = {295},
    publisher = {Elsevier Ltd},
    url = {https://doi.org/10.1016/j.apenergy.2021.117061},
    doi = {10.1016/j.apenergy.2021.117061},
    issn = {03062619}
}

@article{Voyant2017MachineReview,
    title = {{Machine learning methods for solar radiation forecasting: A review}},
    year = {2017},
    journal = {Renewable Energy},
    author = {Voyant, Cyril and Notton, Gilles and Kalogirou, Soteris and Nivet, Marie Laure and Paoli, Christophe and Motte, Fabrice and Fouilloy, Alexis},
    pages = {569--582},
    volume = {105},
    publisher = {Elsevier Ltd},
    url = {http://dx.doi.org/10.1016/j.renene.2016.12.095},
    doi = {10.1016/j.renene.2016.12.095},
    issn = {18790682}
}

@article{Nespoli2022MachineImagery,
    title = {{Machine Learning techniques for solar irradiation nowcasting: Cloud type classification forecast through satellite data and imagery}},
    year = {2022},
    journal = {Applied Energy},
    author = {Nespoli, Alfredo and Niccolai, Alessandro and Ogliari, Emanuele and Perego, Giovanni and Collino, Elena and Ronzio, Dario},
    number = {April 2021},
    pages = {117834},
    volume = {305},
    publisher = {Elsevier Ltd},
    url = {https://doi.org/10.1016/j.apenergy.2021.117834},
    doi = {10.1016/j.apenergy.2021.117834},
    issn = {03062619}
}

@misc{Bergstra2013MakingArchitectures,
    title = {{Making a Science of Model Search: Hyperparameter Optimization in Hundreds of Dimensions for Vision Architectures}},
    year = {2013},
    author = {Bergstra, James and Yamins, Daniel and Cox, David},
    month = {2},
    pages = {115--123},
    volume = {28},
    publisher = {PMLR},
    url = {https://proceedings.mlr.press/v28/bergstra13.html},
    issn = {1938-7228}
}

@article{Visser2022OperationalDistribution,
    title = {{Operational day-ahead solar power forecasting for aggregated PV systems with a varying spatial distribution}},
    year = {2022},
    journal = {Renewable Energy},
    author = {Visser, Lennard and AlSkaif, Tarek and van Sark, Wilfried},
    pages = {267--282},
    volume = {183},
    publisher = {Elsevier Ltd},
    url = {https://doi.org/10.1016/j.renene.2021.10.102},
    doi = {10.1016/j.renene.2021.10.102},
    issn = {18790682}
}

@article{Yang2019OperationalMarket,
    title = {{Operational solar forecasting for the real-time market}},
    year = {2019},
    journal = {International Journal of Forecasting},
    author = {Yang, Dazhi and Wu, Elynn and Kleissl, Jan},
    number = {4},
    pages = {1499--1519},
    volume = {35},
    publisher = {Elsevier B.V.},
    url = {https://doi.org/10.1016/j.ijforecast.2019.03.009},
    doi = {10.1016/j.ijforecast.2019.03.009},
    issn = {01692070}
}

@article{Si2021PhotovoltaicPosition,
    title = {{Photovoltaic power forecast based on satellite images considering effects of solar position}},
    year = {2021},
    journal = {Applied Energy},
    author = {Si, Zhiyuan and Yang, Ming and Yu, Yixiao and Ding, Tingting},
    number = {May},
    pages = {117514},
    volume = {302},
    publisher = {Elsevier Ltd},
    url = {https://doi.org/10.1016/j.apenergy.2021.117514},
    doi = {10.1016/j.apenergy.2021.117514},
    issn = {03062619}
}

@article{Wang2022PredRNN:Learning,
    title = {{PredRNN: A Recurrent Neural Network for Spatiotemporal Predictive Learning}},
    year = {2022},
    journal = {IEEE Transactions on Pattern Analysis and Machine Intelligence},
    author = {Wang, Yunbo and Wu, Haixu and Zhang, Jianjin and Gao, Zhifeng and Wang, Jianmin and Yu, Philip S. and Long, Mingsheng},
    number = {},
    pages = {1--16},
    volume = {},
    url = {http://arxiv.org/abs/2103.09504},
    doi = {10.1109/TPAMI.2022.3165153},
    arxivId = {2103.09504}
}

@article{Wang2017PredRNN:LSTMs,
    title = {{PredRNN: Recurrent neural networks for predictive learning using spatiotemporal LSTMs}},
    year = {2017},
    journal = {Advances in Neural Information Processing Systems},
    author = {Wang, Yunbo and Long, Mingsheng and Wang, Jianmin and Gao, Zhifeng and Yu, Philip S.},
    number = {Nips},
    pages = {880--889},
    volume = {2017-Decem},
    issn = {10495258}
}

@article{Hou2019ProbabilisticChina,
    title = {{Probabilistic duck curve in high PV penetration power system: Concept, modeling, and empirical analysis in China}},
    year = {2019},
    journal = {Applied Energy},
    author = {Hou, Qingchun and Zhang, Ning and Du, Ershun and Miao, Miao and Peng, Fei and Kang, Chongqing},
    month = {5},
    pages = {205--215},
    volume = {242},
    publisher = {Elsevier Ltd},
    doi = {10.1016/J.APENERGY.2019.03.067},
    issn = {03062619}
}

@article{Lin2022RecentMethods,
    title = {{Recent advances in intra-hour solar forecasting: A review of ground-based sky image methods}},
    year = {2022},
    journal = {International Journal of Forecasting},
    author = {Lin, Fan and Zhang, Yao and Wang, Jianxue},
    number = {},
    publisher = {Elsevier B.V.},
    url = {https://doi.org/10.1016/j.ijforecast.2021.11.002},
    doi = {10.1016/j.ijforecast.2021.11.002},
    issn = {01692070}
}

@article{Hewamalage2021RecurrentDirections,
    title = {{Recurrent Neural Networks for Time Series Forecasting: Current status and future directions}},
    year = {2021},
    journal = {International Journal of Forecasting},
    author = {Hewamalage, Hansika and Bergmeir, Christoph and Bandara, Kasun},
    number = {1},
    pages = {388--427},
    volume = {37},
    doi = {10.1016/j.ijforecast.2020.06.008},
    issn = {01692070},
    arxivId = {1909.00590}
}

@article{Antonanzas2016ReviewForecasting,
    title = {{Review of photovoltaic power forecasting}},
    year = {2016},
    journal = {Solar Energy},
    author = {Antonanzas, J. and Osorio, N. and Escobar, R. and Urraca, R. and Martinez-de-Pison, F. J. and Antonanzas-Torres, F.},
    pages = {78--111},
    volume = {136},
    publisher = {Elsevier Ltd},
    url = {http://dx.doi.org/10.1016/j.solener.2016.06.069},
    doi = {10.1016/j.solener.2016.06.069},
    issn = {0038092X}
}

@article{Lin2020Self-attentionPrediction,
    title = {{Self-attention ConvLSTM for spatiotemporal prediction}},
    year = {2020},
    journal = {AAAI 2020 - 34th AAAI Conference on Artificial Intelligence},
    author = {Lin, Zhihui and Li, Maomao and Zheng, Zhuobin and Cheng, Yangyang and Yuan, Chun},
    pages = {11531--11538},
    isbn = {9781577358350},
    doi = {10.1609/aaai.v34i07.6819},
    issn = {2159-5399}
}

@article{duPlessis2021Short-termBehaviour,
    title = {{Short-term solar power forecasting: Investigating the ability of deep learning models to capture low-level utility-scale Photovoltaic system behaviour}},
    year = {2021},
    journal = {Applied Energy},
    author = {du Plessis, A. A. and Strauss, J. M. and Rix, A. J.},
    number = {December 2020},
    pages = {116395},
    volume = {285},
    publisher = {Elsevier Ltd},
    url = {https://doi.org/10.1016/j.apenergy.2020.116395},
    doi = {10.1016/j.apenergy.2020.116395},
    issn = {03062619}
}

@article{Cheng2022Short-termInterest,
    title = {{Short-term Solar Power Prediction Learning Directly from Satellite Images with Regions of Interest}},
    year = {2022},
    journal = {IEEE Transactions on Sustainable Energy},
    author = {Cheng, Lilin and Zang, Haixiang and Wei, Zhinong and Ding, Tao and Xu, Ruiqi and Sun, Guoqiang},
    number = {1},
    month = {1},
    pages = {629--639},
    volume = {13},
    publisher = {Institute of Electrical and Electronics Engineers Inc.},
    doi = {10.1109/TSTE.2021.3123476},
    issn = {19493037}
}

@article{Fu2021SkyForecasting,
    title = {{Sky Image Prediction Model Based on Convolutional Auto-Encoder for Minutely Solar PV Power Forecasting}},
    year = {2021},
    journal = {IEEE Transactions on Industry Applications},
    author = {Fu, Yuwei and Chai, Hua and Zhen, Zhao and Wang, Fei and Xu, Xunjian and Li, Kangping and Shafie-Khah, Miadreza and Dehghanian, Payman and Catalao, Joao P.S.},
    number = {4},
    pages = {3272--3281},
    volume = {57},
    doi = {10.1109/TIA.2021.3072025},
    issn = {19399367}
}

@article{Fouilloy2018SolarVariability,
    title = {{Solar irradiation prediction with machine learning: Forecasting models selection method depending on weather variability}},
    year = {2018},
    journal = {Energy},
    author = {Fouilloy, Alexis and Voyant, Cyril and Notton, Gilles and Motte, Fabrice and Paoli, Christophe and Nivet, Marie Laure and Guillot, Emmanuel and Duchaud, Jean Laurent},
    pages = {620--629},
    volume = {165},
    url = {https://doi.org/10.1016/j.energy.2018.09.116},
    doi = {10.1016/j.energy.2018.09.116},
    issn = {03605442}
}

@article{Sobri2018SolarReview,
    title = {{Solar photovoltaic generation forecasting methods: A review}},
    year = {2018},
    journal = {Energy Conversion and Management},
    author = {Sobri, Sobrina and Koohi-Kamali, Sam and Rahim, Nasrudin Abd},
    number = {December 2017},
    pages = {459--497},
    volume = {156},
    publisher = {Elsevier},
    url = {https://doi.org/10.1016/j.enconman.2017.11.019},
    doi = {10.1016/j.enconman.2017.11.019},
    issn = {01968904}
}

@article{Yang2020VerificationForecasts,
    title = {{Verification of deterministic solar forecasts}},
    year = {2020},
    journal = {Solar Energy},
    author = {Yang, Dazhi and Alessandrini, Stefano and Antonanzas, Javier and Antonanzas-Torres, Fernando and Badescu, Viorel and Beyer, Hans Georg and Blaga, Robert and Boland, John and Bright, Jamie M. and Coimbra, Carlos F.M. and David, Mathieu and Frimane, Âzeddine and Gueymard, Christian A. and Hong, Tao and Kay, Merlinde J. and Killinger, Sven and Kleissl, Jan and Lauret, Philippe and Lorenz, Elke and van der Meer, Dennis and Paulescu, Marius and Perez, Richard and Perpi{\~{n}}{\'{a}}n-Lamigueiro, Oscar and Peters, Ian Marius and Reikard, Gordon and Renn{\'{e}}, David and Saint-Drenan, Yves Marie and Shuai, Yong and Urraca, Ruben and Verbois, Hadrien and Vignola, Frank and Voyant, Cyril and Zhang, Jie},
    number = {February},
    pages = {20--37},
    volume = {210},
    publisher = {Elsevier},
    url = {https://doi.org/10.1016/j.solener.2020.04.019},
    doi = {10.1016/j.solener.2020.04.019},
    issn = {0038092X}
}

@article{Barbieri2017VeryReview,
    title = {{Very short-term photovoltaic power forecasting with cloud modeling: A review}},
    year = {2017},
    journal = {Renewable and Sustainable Energy Reviews},
    author = {Barbieri, Florian and Rajakaruna, Sumedha and Ghosh, Arindam},
    pages = {242--263},
    volume = {75},
    publisher = {Elsevier Ltd},
    url = {http://dx.doi.org/10.1016/j.rser.2016.10.068},
    doi = {10.1016/j.rser.2016.10.068},
    issn = {18790690}
}

@article{Obeso2022VisualDetection,
    title = {{Visual vs internal attention mechanisms in deep neural networks for image classification and object detection}},
    year = {2022},
    journal = {Pattern Recognition},
    author = {Obeso, Abraham Montoya and Benois-Pineau, Jenny and Garc{\'{i}}a V{\'{a}}zquez, Mireya Saraí and Acosta, Alejandro Álvaro Ramírez},
    pages = {108411},
    volume = {123},
    publisher = {Elsevier Ltd},
    url = {https://doi.org/10.1016/j.patcog.2021.108411},
    doi = {10.1016/j.patcog.2021.108411},
    issn = {00313203}
}

@techreport{IEA2021World2021,
    title = {{World Energy Outlook 2021}},
    year = {2021},
    booktitle = {World Energy Outlook 2021},
    author = {{IEA}},
    pages = {1--386},
    url = {https://www.iea.org/reports/world-energy-outlook-2021},
    address = {Paris}
}

@article { CloudCast,
      author = "Mikko Partio and Leila Hieta and Anniina Kokkonen",
      title = "CloudCast—Total Cloud Cover Nowcasting with Machine Learning",
      journal = "Artificial Intelligence for the Earth Systems",
      year = "2025",
      publisher = "American Meteorological Society",
      address = "Boston MA, USA",
      volume = "4",
      number = "3",
      doi = "10.1175/AIES-D-24-0104.1",
      pages=      "e240104",
      url = "https://journals.ametsoc.org/view/journals/aies/4/3/AIES-D-24-0104.1.xml"
}

@article{ZHOOLID,
title = {Deep learning-based identification of precipitation clouds from all-sky camera data for observatory safety},
journal = {Machine Learning with Applications},
volume = {20},
pages = {100640},
year = {2025},
issn = {2666-8270},
doi = {https://doi.org/10.1016/j.mlwa.2025.100640},
url = {https://www.sciencedirect.com/science/article/pii/S2666827025000234},
author = {Mohammad H. {Zhoolideh Haghighi} and Alireza Ghasrimanesh and Habib Khosroshahi},
}

@article{HOU2025126243,
title = {Effects of clouds, aerosols and shadows on solar energy potential on urban rooftops using earth observation data and digital surface models},
journal = {Applied Energy},
volume = {397},
pages = {126243},
year = {2025},
issn = {0306-2619},
doi = {https://doi.org/10.1016/j.apenergy.2025.126243},
url = {https://www.sciencedirect.com/science/article/pii/S0306261925009730},
author = {Xinyuan Hou and Kyriakoula Papachristopoulou and Yves-Marie Saint-Drenan and Panagiotis G. Kosmopoulos and Basil E. Psiloglou and Stelios Kazadzis},
}

@article{BARANCSUK2025122962,
title = {Hybrid ultra-short term solar irradiation forecasting using resource-efficient multi-step long-short term memory},
journal = {Renewable Energy},
volume = {247},
pages = {122962},
year = {2025},
issn = {0960-1481},
doi = {https://doi.org/10.1016/j.renene.2025.122962},
url = {https://www.sciencedirect.com/science/article/pii/S096014812500624X},
author = {Lilla Barancsuk and Veronika Groma and Barnabás Kocziha},}

@article{perera2024day,
  title={Day-ahead regional solar power forecasting with hierarchical temporal convolutional neural networks using historical power generation and weather data},
  author={Perera, Maneesha and De Hoog, Julian and Bandara, Kasun and Senanayake, Damith and Halgamuge, Saman},
  journal={Applied Energy},
  volume={361},
  pages={122971},
  year={2024},
  publisher={Elsevier}
}

@article{perera2022multi,
  title={Multi-resolution, multi-horizon distributed solar PV power forecasting with forecast combinations},
  author={Perera, Maneesha and De Hoog, Julian and Bandara, Kasun and Halgamuge, Saman},
  journal={Expert Systems with Applications},
  volume={205},
  pages={117690},
  year={2022},
  publisher={Elsevier}
}

@article{wang2025multimodal,
  title={Multimodal Ensemble Photovoltaic Power Forecasting Incorporating Diffusion-based Satellite Image Prediction},
  author={Wang, Kai and Wang, Tao and Shan, Shuo and Dou, Weijing and Zhang, Jingxin and Wei, Haikun and Zhang, Kanjian},
  journal={IEEE Transactions on Sustainable Energy},
  year={2025},
  publisher={IEEE}
}
